%% file: main.tex
\documentclass[letterpaper, 10 pt, conference, usenames, dvipsnames]{ieeeconf}  

\IEEEoverridecommandlockouts                              

\usepackage[dvipdfmx]{graphicx}
\graphicspath{{figs/}}
\usepackage{amsmath}
\usepackage{amssymb}
\usepackage{algorithm}
\usepackage{algpseudocode}
\usepackage{times}
\usepackage[position=bottom]{subfig}
\usepackage{xcolor}
\usepackage{hyperref}
\hypersetup{colorlinks=true}
\usepackage[space]{cite}
\usepackage[justification=centering]{caption}
\usepackage{multirow}
\newcommand{\figlab}[1]{\label{figure:#1}}
\newcommand{\tablab}[1]{\label{table:#1}}
\newcommand{\figref}[1]{Figure \ref{figure:#1}}
\newcommand{\tabref}[1]{Table \ref{table:#1}}
\newcommand{\secref}[1]{$\S$\ref{section:#1}}

\title{\LARGE \bf
Instance Segmentation of Visible and Occluded Regions \\
for Finding and Picking Target from a Pile of Objects
}

\author{Kentaro Wada, Shingo Kitagawa, Kei Okada and Masayuki Inaba
  \\
  University of Tokyo, JSK Laboratory \\
  \{wada, s-kitagawa, okada, inaba\}@jsk.imi.i.u-tokyo.ac.jp%
}

\begin{document}

\maketitle
\thispagestyle{empty}
\pagestyle{empty}

\begin{abstract}
\input src/abstract.tex
\end{abstract}

\input src/introduction.tex
\input src/related_works.tex
\input src/methods.tex
\input src/experiments.tex
\input src/conclusions.tex
\input src/acknowledgement.tex



\bibliographystyle{junsrt}
\bibliography{main}

\end{document}

%% file: src/abstract.tex
We present a robotic system for picking a target from a pile of objects that is capable of finding and grasping the target object by removing obstacles in the appropriate order. The fundamental idea is to segment instances with both visible and occluded masks, which we call `instance occlusion segmentation'. To achieve this, we extend an existing instance segmentation model with a novel `relook' architecture, in which the model explicitly learns the inter-instance relationship. Also, by using image synthesis, we make the system capable of handling new objects without human annotations. The experimental results show the effectiveness of the relook architecture when compared with a conventional model and of the image synthesis when compared to a human-annotated dataset. We also demonstrate the capability of our system to achieve picking a target in a cluttered environment with a real robot.

%% file: src/introduction.tex
\section{INTRODUCTION}

With recent progress in deep learning, especially convolutional neural networks, the robotics community has improved the ability of the robot to find and pick various target objects in clutter~\cite{Jonschkowski2016,Hernandez2017,Schwarz2017,Zeng2017,wada2017probabilistic}, even when including novel objects~\cite{Zeng2018}.  However, these works restrict the environment to have few occluded target objects.  Our goal in this work is to develop a framework that enables a robot to pick various target objects in the appropriate order in an environment with heavy occlusion (e.g., a pile of objects in a bin).

\begin{figure*}[t]
  \centering
  \includegraphics[width=0.9\textwidth]{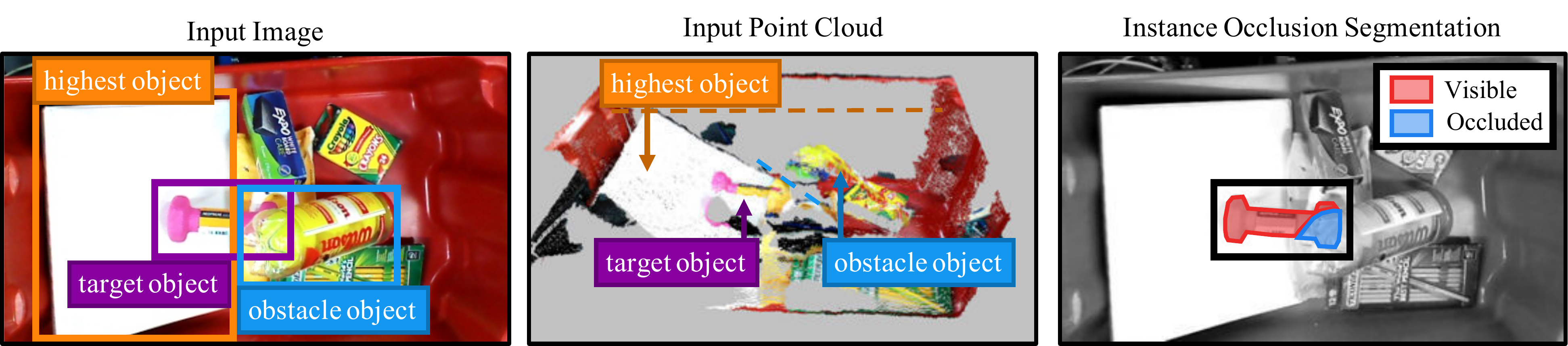}
  \caption{\textbf{Typical case where occlusion understanding is necessary.}
    \\
    \footnotesize{
      The target object (\textcolor{Plum}{Dumbbell})
      is located on top of the highest object (\textcolor{RedOrange}{Binder})
      and occluded by the obstacle object (\textcolor{ProcessBlue}{Tennis Ball Holder}).\\
      In this case, we expect the robot to pick the Tennis Ball Holder first and the target object afterward.
      However, the robot mistakenly tries to pick the Binder first\\
      in height order picking or the Dumbbell in greedy picking.
    }
  }
  \label{figure:problem}
\end{figure*}

Picking target objects from a pile is especially difficult with a vast variety of shapes and arrangement of deformable objects.  Estimating the state of stacked objects is challenging because of 1) Partial observation for the mesh model fitting; 2) Infinite possible patterns of stacking.
For example, in \figref{problem}, the target object (\textcolor{Plum}{Dumbbell}) is located on the highest object (\textcolor{RedOrange}{Binder}) and under the obstacle object (\textcolor{ProcessBlue}{Tennis Ball Holder}).  In this scenario, we expect the robot to pick the Tennis Ball Holder first and the Dumbbell afterward, without touching any other objects. However, the detection of only the visible part is not enough to plan such a path to grasp the target object.
Typical failure cases occur when the robot tries to pick the wrong object first: 1) Binder because it is the highest object; 2) Dumbbell because it is the target object (\figref{problem}).

In order to plan the picking order correctly in this situation, it is necessary to understand the scene as ``{\it target object Dumbbell is occluded by obstacle object Tennis Ball Holder}''.  This motivates us to introduce {\it instance occlusion segmentation}, in which the occluded region of each instance (i.e. object) is segmented as well as the visible one.
The segmentation task is aimed at predicting the invisible information (e.g., occluded region) from the visible one using some external knowledge (e.g., by learning from a dataset).  This spontaneously leads us to applying a learning-based method as the solution to this problem.

\begin{figure}[t]
  \centering
  \includegraphics[width=\columnwidth]{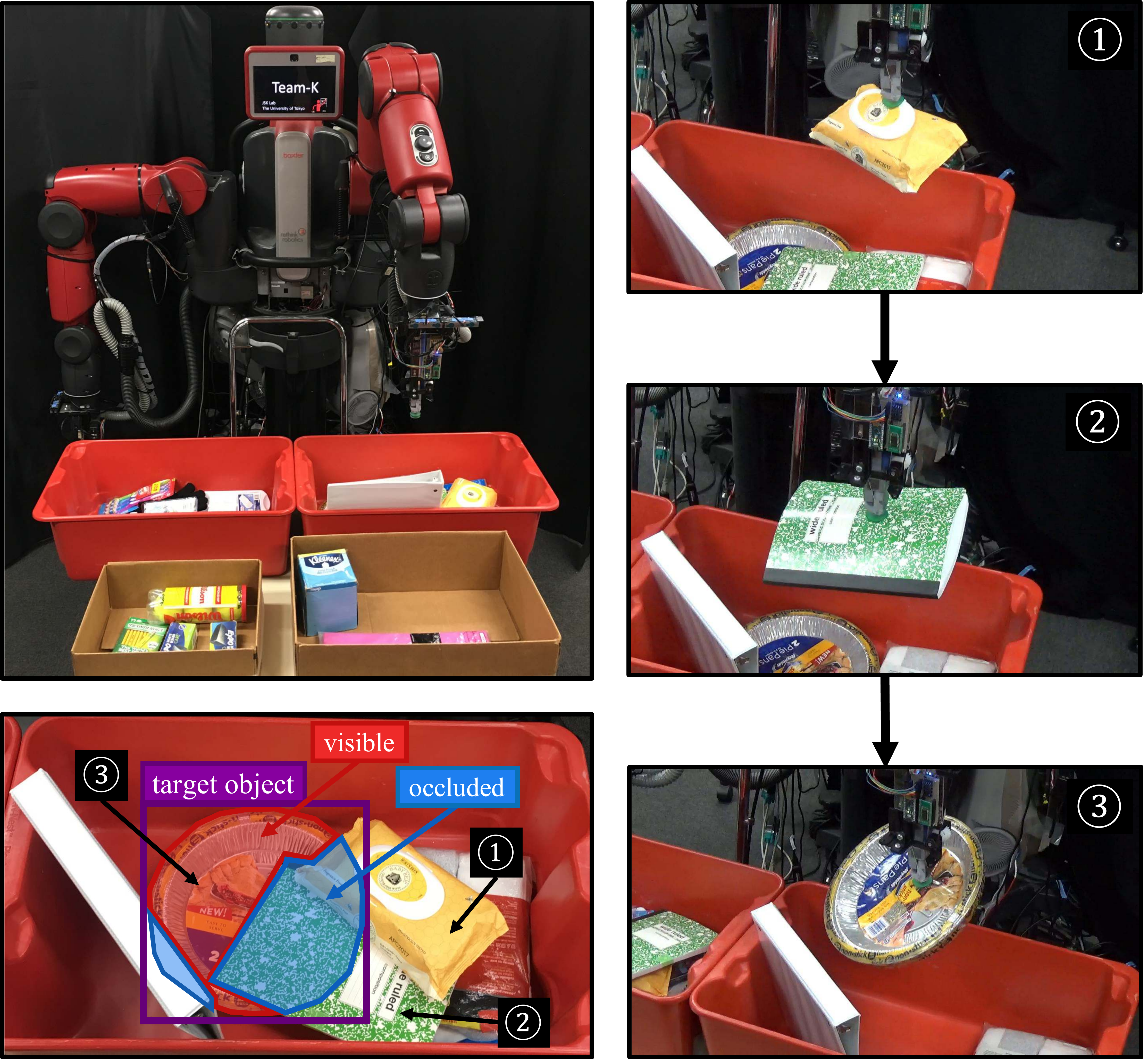}
  \caption{{\bf Our system} recognizes occlusion status of objects
    via instance occlusion segmentation, and plans an appropriate picking order
    to pick the target object from the stacked objects.}
  \label{figure:overview}
\end{figure}

In this paper, we propose a vision system that detects objects and recognizes their visible and occluded region simultaneously.
The system is designed to be able to handle novel objects without requiring any task-specific training data (e.g., data collection and human annotation for objects in the bin).  To achieve this, our system consists of two components: 1) Image synthesis of a stack of objects solely composed by instance images of these objects, which generates various patterns of stacking and occlusion with ground truth of visible and occluded region masks for each instance; 2) Instance segmentation model with a novel `relook' architecture designed for occlusion segmentation, in which we extend the recent works~\cite{He2017,Do2017} to recognize and use the density of instances for multi-class segmentation (visible and occluded) of each instance.  We also propose a metric for instance occlusion segmentation, which is an extension of common instance segmentation which only segments the visible mask.  In the experiments, we provide evaluation results of instance occlusion segmentation using various objects that were used in the Amazon Robotics Challenge (ARC), demonstrating the ability of our system to find and pick the target from a pile of objects.

Our main contributions are:
\begin{itemize}
  \item Image synthesis framework for learning instance segmentation
        including occlusion, which is a straightforward extension to recent works~\cite{Georgakis2017,Dwibedi2017};
  \item A novel instance segmentation network model that uses the instance density
        to segment multi-class masks by extending recent works~\cite{He2017,Do2017};
  \item A new metric for instance segmentation of multi-class masks extending recent work
        of instance segmentation of a single mask~\cite{Kirillov2018};
  \item The integrated system of the above components and demonstration
        in the picking task from a pile of objects.
\end{itemize}

%% file: src/related_works.tex
\section{RELATED WORKS}

\subsection{Instance Segmentation}\label{section:related_works:instance_seg}

Instance segmentation is aimed at predicting the object region mask and its label at the same time.  Since instance segmentation is a compound task of bounding box detection and pixel-wise semantic segmentation, previous works propose models that solve the two tasks {\it sequentially} or {\it concurrently}.  In the {\it sequential} approach, past works~\cite{Pinheiro2015,Pinheiro2016,Dai2016} propose models which propose mask segmentation first and classifies them afterward.  On the other hand, {\it concurrent} segmentation prediction and object detection are recently proposed~\cite{Li2016,He2017,Do2017}.  These models simultaneously predict object classes, boxes, and masks.  In fact, these concurrent approaches prove to be faster and more accurate than sequential ones.

In this paper, we extend the state-of-the-art work~\cite{He2017,Do2017}, for instance occlusion segmentation by seeing it as the multi-class (visible, occluded) extension of the conventional instance segmentation (visible only).  Compared to the visible region segmentation, the relationship between nearby objects is more crucial when dealing with occlusion segmentation, since the occluded region of an instance is caused by other instance's visible region.  This motivates us to extend the previous models in order to learn the connection between predicted object instances. Although previous models concurrently predict visible region masks for each box, there is no connection between them.

Specifically, multi-class extension of {\it Mask R-CNN}~\cite{He2017} is proposed as {\it AffordanceNet}~\cite{Do2017}.  It replaces sigmoid cross entropy loss with softmax to output multi-class region masks, so we refer to the {\it AffordanceNet} as {\it Mask R-CNN Softmax} later in this paper.


\subsection{Image Synthesis for Object Detection}

Recently deep learning based methods have improved many machine vision tasks, but since deep learning requires a significant amount of data, it motivates researchers to acquire training data through synthesizing rather than using human annotations.
A naive approach of synthesizing training data is using 3D mesh models to render 2D images.  Past works use mesh models to learn viewpoint estimation~\cite{Su2015}, eye gaze direction estimation~\cite{Shrivastava2016} and object detection~\cite{Hinterstoisser2017} for objects in a 2D image.  Above work focus on developing ways to make synthetic images closer to real images.  On the other hand, it has recently been found that synthesizing only 2D instance images of objects is also effective to train detection models of object bounding boxes~\cite{Georgakis2017,Dwibedi2017}.  The base idea for this is that if we could generate infinite synthetic images at random and train learning model with it, the model would generalize to real images.  Our approach is closer to the latter, and we extend the past works to generate ground truth of object masks (visible, occluded) as well, in addition to the bounding box.  Since it is impractical to gather realistic mesh models for various objects, 2D synthesizing is more practical than 3D one.  We also show that a small number of instance images is enough to achieve human annotation-level detection performance, while past works~\cite{Georgakis2017,Dwibedi2017} use a huge number of instance images for each object.



%% file: src/methods.tex
\section{System Overview}

\begin{figure*}[htbp]
  \centering
  \includegraphics[width=\textwidth]{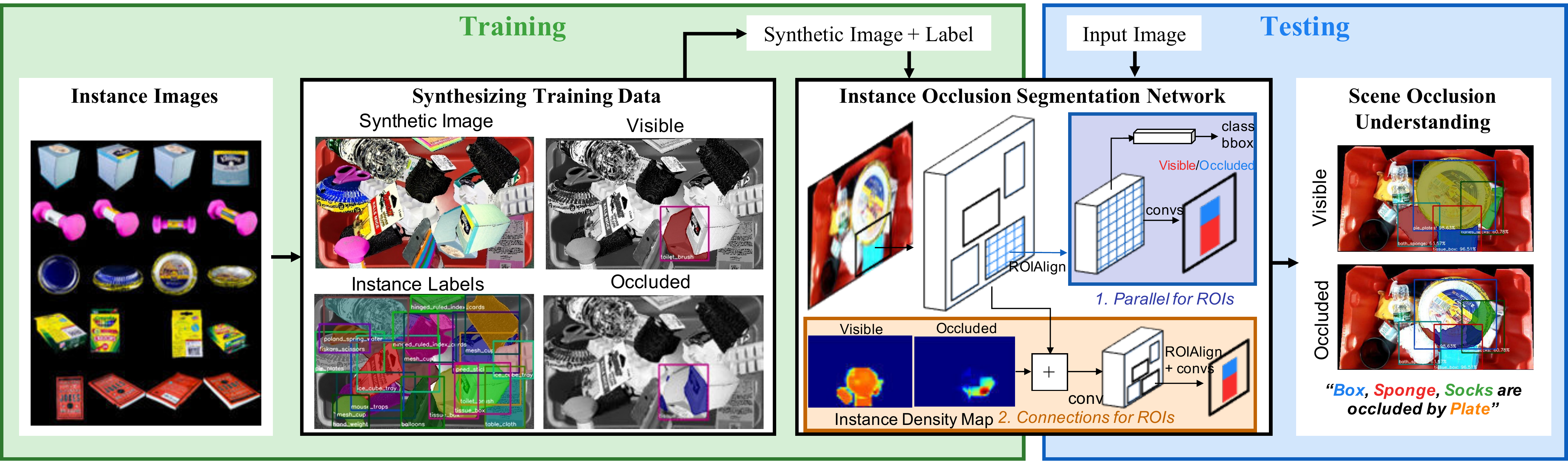}
  \caption{\textbf{System overview.}\\
    \footnotesize{%
      At \textcolor{Green}{train} time, the system receives instance images as input
      and trains the segmentation model.\\
      At \textcolor{blue}{test} time, the system receives a real image as input
      and outputs the scene occlusion status for the image.
    }
  }
  \figlab{system}
\end{figure*}

Our proposed system (\figref{system}) is composed of two components:
\begin{itemize}
  \item Instance occlusion segmentation neural networks trained using the generated images
        (\secref{instance_seg_nn});
  \item 2D image synthesis of a cluttered scene generated from object instance images
        to handle various objects (\secref{image_synthesis}).
\end{itemize}
At training time, the system receives instance images of objects as the input to train instance occlusion segmentation model from the generated synthetic images.  At testing time, a real image is the input and the occlusion status of the scene is predicted and outputted. Since our proposed system only requires instance images of objects of interest, we can quickly gather them from web or with a standard camera. This is important for enabling the vision system to handle various objects without human labeling, which is especially hard for instance occlusion segmentation, and for the applications to warehouse picking in e-commerce services for example.

The network output is the set of visible and occluded masks of each object instance and the occlusion status. The hierarchy of stacking like ``{\it Plate is on Box\/}`` is interpreted automatically by scanning each pixel to see what object is visible and what objects are occluded at the pixel.


\section{Instance Occlusion Segmentation Neural Networks}\label{section:instance_seg_nn}

\subsection{Network Architecture}

\subsubsection{Mask R-CNN}

We begin by reviewing the network architecture of Mask R-CNN~\cite{He2017}. Mask R-CNN is an extension of Faster R-CNN~\cite{Ren2017}, which was previously proposed as a model for bounding box based object detection.  To extend the model for instance segmentation, another branch of predicting instance mask was introduced and added to the existing class and bounding box prediction branches.  In the mask branch, the instance mask is predicted without predicting object class by relying on the classification of the classification branch. This is achieved by predicting the class number of instance masks in the mask branch, extracting the mask of predicted class by classification branch afterward. Each branch of the object class, box, and mask is predicted in parallel after extracting features by ROIAlign (an extension of ROIPooling~\cite{Ren2017}) at the Region of Interest (ROI) proposed by the preceding Region Proposal Networks (RPN)~\cite{Ren2017}.

Mask R-CNN is fast because the prediction of each instance information is made in {\it parallel\/} and {\it independently\/} using the features extracted by ROIAlign for each proposed ROI.  However, since the prediction about each instance after ROI feature extraction is independent of other ROIs/instances, this model in not suited for leaning the relationship among instances.  Therefore, although effective for instance segmentation of the visible region, in which the relationship among instances is not so important, Mask R-CNN is not expected to be suitable for instance occlusion segmentation, in which the relationship between masks is crucial for correctly inferring the occlusion state of the instances.  This leads us to introduce the inter-instance connection, which is described in \secref{inter_instance_connection}.


\subsubsection{Relook Architecture: Inter-Instance Connection}\label{section:inter_instance_connection}

In order to learn the relationship and dependency between instances, it is necessary to have connections among the representations of each instance in the neural network.  To do this, we convert the instance masks predicted in the first stage (left in \figref{network_architecture}) to a density map (middle in \figref{network_architecture}) and use it to predict instance masks in the second stage.  The two instance masks of the first and second stage are added in pixel-wise (fused) as the final result, being segmentation loss computed for this fused result.  The first stage is Mask R-CNN (softmax) (\secref{related_works:instance_seg}). The second stage can be interpreted as a ``{\it relook}'' architecture for learning and predicting inter-instance connections, so we call this model Mask R-CNN (relook).

\begin{figure*}[htbp]
  \centering
  \includegraphics[width=0.95\textwidth]{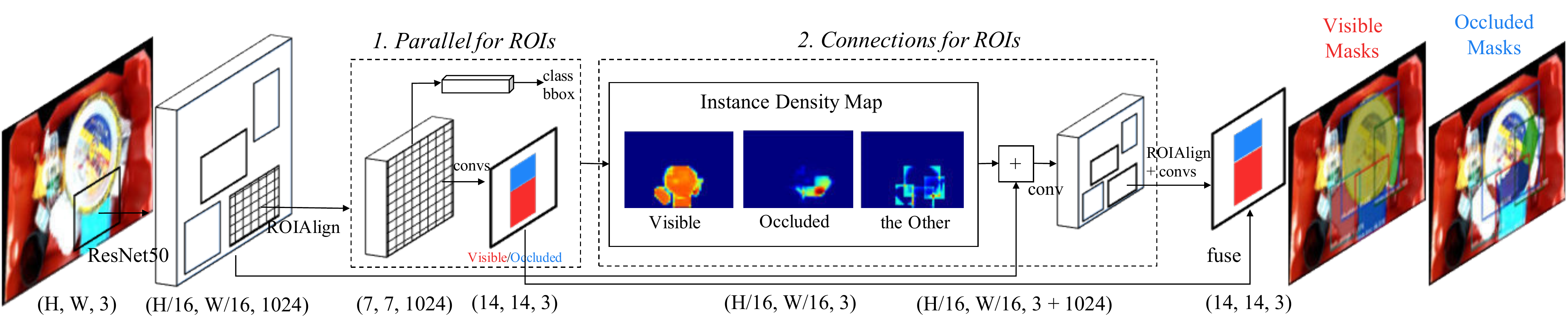}
  \caption{\textbf{Instance occlusion segmentation network with inter-instance connection.}\\
    \footnotesize{%
      The network architecture can be interpreted as a pipeline of two components:\\
      1) Mask R-CNN (softmax) for per-instance prediction;
      2) Instance density map for inter-instance connection.\\
      The tensor shape of the layer output is shown in the bottom with (height, width, channel).
    }
  }
  \figlab{network_architecture}
\end{figure*}

The final layer predicts three masks: visible, occluded and other (not part of the object), being the density map also generated for each of the three.  For learning inter-instance connection, we concatenate the density map with the features extracted by the feature extractor. We use ResNet50-C4 and ResNet101-C4 (fourth stage output of ResNetX~\cite{Nair2010}) as in Mask R-CNN~\cite{He2017}, and apply a convolutional layer.  After that, ROIAlign, `res5' (5th layer of ResNetX) and a deconvolutional layer is applied with sharing parameters with the first stage.  On top of that, the final convolutional layer is applied to predict three masks for each instance.  We use ReLU~\cite{Nair2010} as the activation function for the hidden layers, and the kernel size of the convolutional layer is 3 for the hidden layer and 1 for the output layer.


\subsection{Implementation Detail}

At most, we followed the implementation of the original paper Mask R-CNN~\cite{He2017} and Faster R-CNN~\cite{Ren2017}.  For input image size, we set 600 as the minimum axis size and 1000 as the maximum axis size following~\cite{Ren2017}.  We trained RPN using three ratios and four anchor scales with no threshold of ROI proposal size, which was set to 16 in~\cite{Ren2017}. We use 512 as the hidden channel size of RPN, value which was set to 1024 in~\cite{He2017}, since we noticed it was too large for small datasets.

At the training time of Mask R-CNN, 512 ROI proposals, which have $foreground:background = 1:4$ ratio, are used to train classification and box regression branch, and the foreground ROIs are used to train the mask branch. At the testing time, it first regresses class and bounding box of the instances and applies non-maximum suppression (NMS), then using the fewer and more accurate bounding boxes for ROI feature extraction by ROIAlign with following mask prediction.  This causes a little difference in the prediction process between training and testing; however, it achieves mask prediction using the more accurate bounding box to get better results.  This prediction difference may also cause bad effects because of the difference of the density map: at training time NMS is not applied, so the density map appears to have more instances than that of testing time.  However, a little surprisingly, this prediction difference did not cause adverse effects, but rather we had better results with this prediction process than using the same prediction process at both training and testing time.


\section{Image Synthesis for Learning Instance Occlusion Segmentation}%
\label{section:image_synthesis}

\subsection{Instance Image Gathering}\label{section:instance_image_gathering}

We use instance images distributed at the Amazon Robotics Challenge as {\it ItemData\/} for both known and novel objects which will appear during the task.  There are few images ($4 - 6$) taken from different viewpoints for each object with black background, as shown in the top images of \figref{instance_images_masks}.


\begin{figure}[htbp]
  \centering
  \includegraphics[width=\columnwidth]{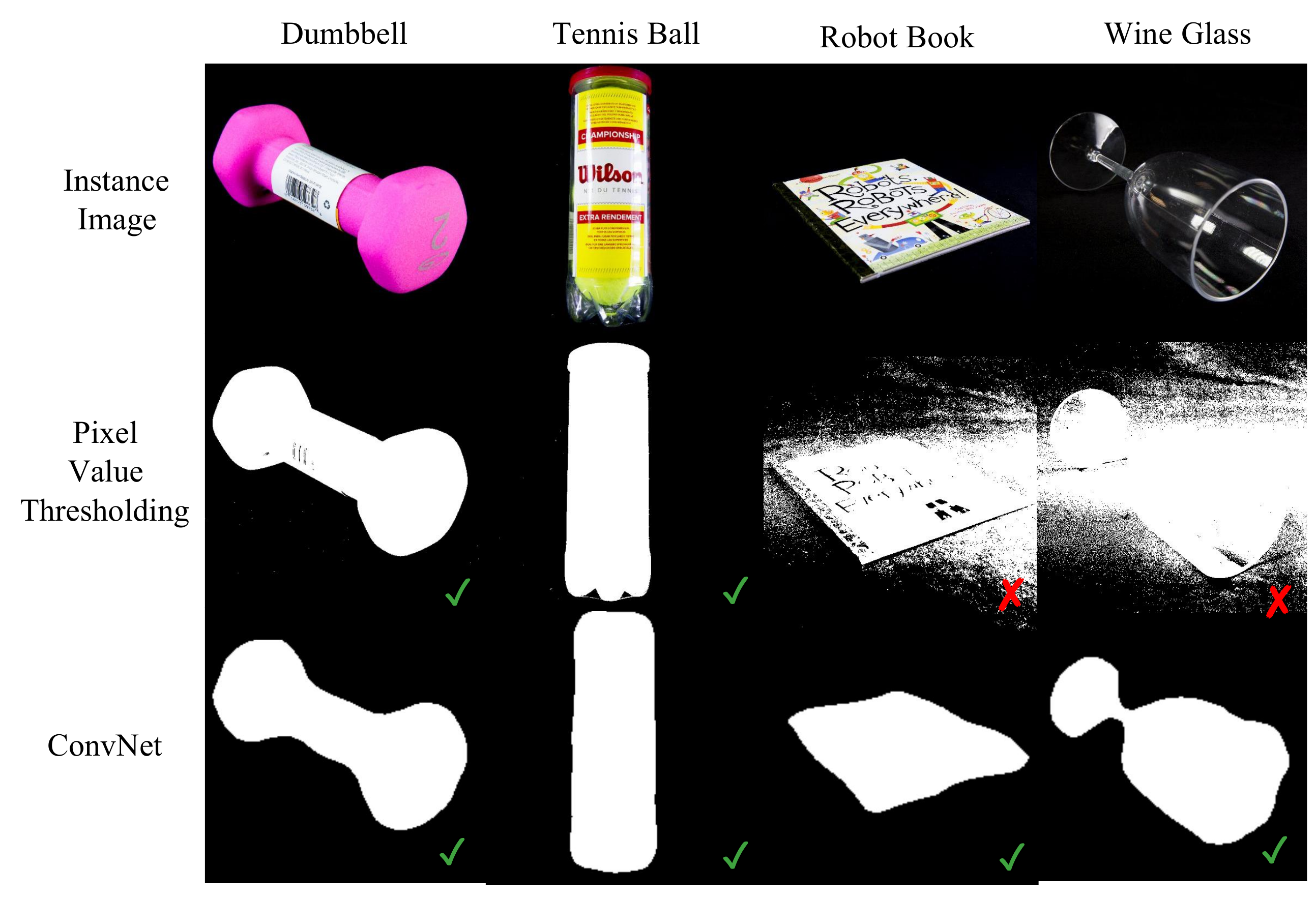}
  \caption{\textbf{Foreground mask extraction for instance image.}\\
    \footnotesize{Sometimes foreground mask extraction through pixel value threshold fails even for images with simple background (middle images).\\
    In this work we counteract this problem by training a small convolutional neural network model (bottom images)~\cite{Long2015}.}
    }
  \figlab{instance_images_masks}
\end{figure}

Although the background of the instance image is static, we found that it is difficult to extract the foreground mask for each instance through pixel value thresholding.  This is especially true for objects with various colors (e.g., Robot Book) or transparent (e.g., Wine Glass), as shown in the middle of \figref{instance_images_masks}.

A similar problem is also pointed out in previous work~\cite{Dwibedi2017}: they had objects for which it is difficult to extract foreground mask by thresholding of the depth image, in particular with transparent objects such as Cola Bottle.  To overcome this difficulty, they trained a small convolutional neural network (ConvNet) model~\cite{Long2015} using the mask acquired from thresholding of the depth image as the ground truth.  We applied the same approach, using the mask acquired from pixel value thresholding using instance images of 112 objects.  The ground truth foreground mask is not perfect because it is not annotated by a human; however, the segmentation result in \figref{instance_images_masks} shows that the network successfully generalized the foreground segmentation.  In this case, the automatically generated mask is usually bigger than the perfect mask, while in~\cite{Dwibedi2017} the generated mask from depth image is usually smaller than the perfect one.

\subsection{Blending}\label{section:blending}

Blending is necessary for 2D image synthesis to remove boundary artifacts when we put the instance images onto the background image.  We apply gaussian blurring following~\cite{Dwibedi2017} with a random sigma which ranges from 0 to 1.

\subsection{Data Augmentation}\label{section:data_augmentation}

Data augmentation is crucial for generating synthetic data to train detection models that will generalize to real images, especially when we only have a few instance images ($4 - 6$), compared to 600 images found in~\cite{Dwibedi2017}.  We applied color data augmentation in addition to the geometric augmentation present in past work~\cite{Dwibedi2017}.  To make the system more robust to changes in brightness and light reflection, we also applied multiplication to S and V channels after converting the RGB image into HSV color space with random selection of scale between 0.5 to 2.0.  For geometric augmentation, we applied affine transformation with a random scale in the range from 0.5 to 1.0, translation from -16 to 16 pixels, rotation from -180 to 180 degrees, and shear from -16 to 16 degrees, which we believe is similar to the augmentation of 2D/3D rotation described in~\cite{Dwibedi2017}.  \figref{data_augmentation} shows some examples of the augmented result.

\begin{figure}[htbp]
  \centering
  \includegraphics[width=0.95\columnwidth]{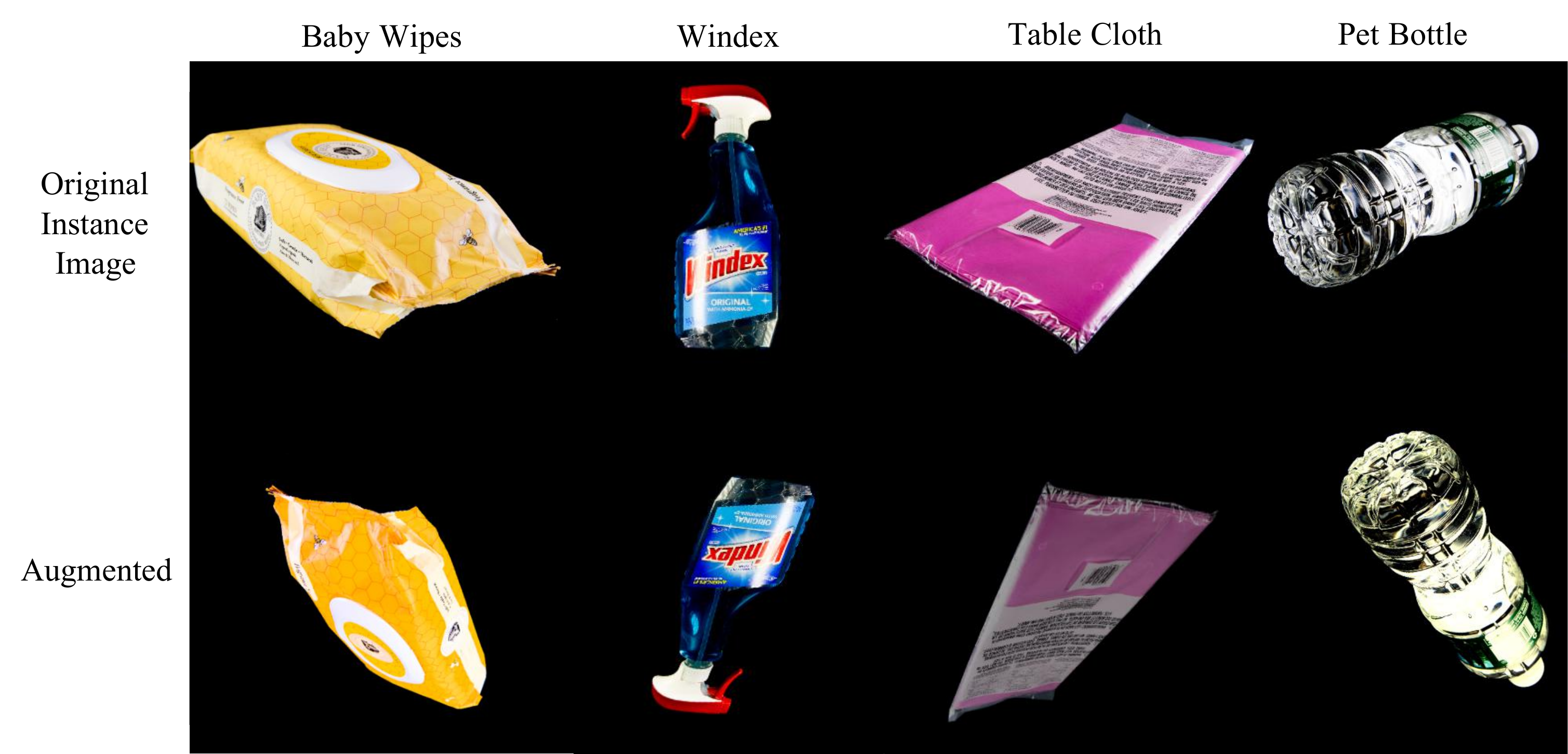}
  \caption{\textbf{Data augmentation example.}}
  \figlab{data_augmentation}
\end{figure}

\subsection{Image Synthesis of a Pile of Objects with Ground Truth}

We generate the ground truth labels and masks in addition to the synthetic image of stacked objects.  \figref{image_synthesis} shows an example of the synthesized image and visualization of the ground truth.  While stacking the instance images onto the background image, we apply image blending (\secref{blending}) and data augmentation (\secref{data_augmentation}) for each instance image at random.  The foreground mask of each instance image is acquired by ConvNet (\secref{instance_image_gathering}), and since the filled region of already stacked instances is known, we can get masks of both visible and occluded regions (\figref{image_synthesis:visible} and \ref{figure:image_synthesis:occluded}).

\begin{figure}[htbp]
  \centering
  \subfloat[\textbf{Synthetic Image.}]{
    \includegraphics[width=0.47\columnwidth]{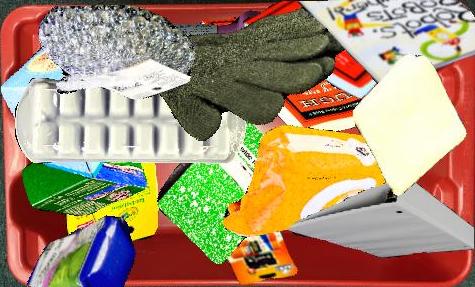}
    \figlab{image_synthesis:image}
  }
  \subfloat[\textbf{Instance Labels.}]{
    \includegraphics[width=0.47\columnwidth]{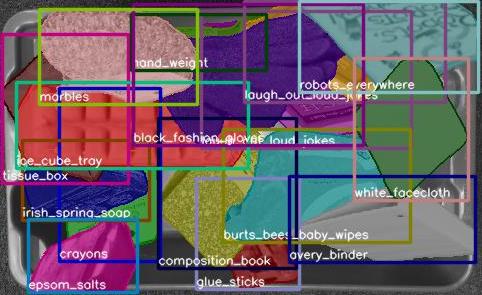}
    \figlab{image_synthesis:labels}
  }

  \subfloat[\textbf{Visible Mask of an Instance.}]{
    \includegraphics[width=0.47\columnwidth]{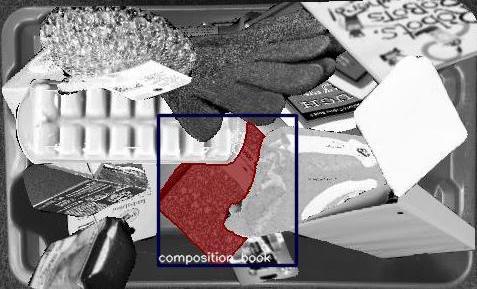}
    \figlab{image_synthesis:visible}
  }
  \subfloat[\textbf{Occluded Mask of an Instance.}]{
    \includegraphics[width=0.47\columnwidth]{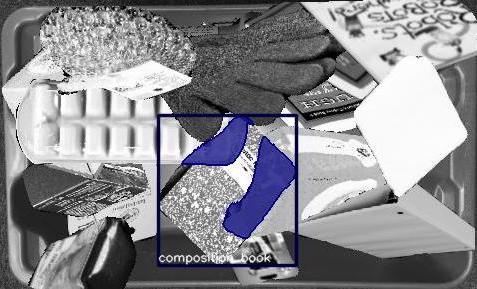}
    \figlab{image_synthesis:occluded}
  }
  \caption{\textbf{Image synthesis example.}}
  \figlab{image_synthesis}
\end{figure}


\section{A METRIC FOR INSTANCE OCCLUSION SEGMENTATION}

The instance segmentation model of multiple masks (affordance masks) in~\cite{Do2017} was evaluated as a pixel-wise segmentation task to compare with other state-of-the-art semantic segmentation models.  This was possible because their model segments only the visible region of each instance and there was no overlap among the masks.  Also, their evaluation ignores the accuracy of the predicted number of instances: even if 3 instances are predicted for an image with 2 instances in ground truth, there is no penalty for the over-counting 1 instance. This happens because the predicted bounding boxes and affordance masks will be converted to an image of pixel-wise affordance label to evaluate it in the semantic segmentation manner.  In order to attain the goal of picking targets from a clutter, however, detecting the correct number of instances is also necessary, as well as the segmentation of visible and occluded masks.  This motivated us to find another metric to evaluate instance occlusion segmentation.

Recently, the metric Panoptic Quality ($PQ$) is proposed to evaluate the accuracy of both detection and segmentation in instance segmentation~\cite{Kirillov2018}.  $PQ$ is represented as the multiplication of Detection Quality ($DQ$) and Segmentation Quality ($SQ$):
\begin{eqnarray}
  DQ &=& \frac{|TP|}{|TP| + \frac{1}{2}|FP| + \frac{1}{2}|FN|} \\
  SQ &=& \frac{\sum_{(p, g) \in TP} IoU(p, q)}{|TP|} \\
  PQ &=& DQ \cdot SQ,
\end{eqnarray}
where $TP$, $FP$ and $FN$ represent the set of true positive, false positive and false negative instances, and $p$, $g$ represent predicted and ground truth instances, respectively.  $IoU$ is the intersect over union of predicted and ground truth masks for a single mask-class:
\begin{eqnarray}
  IoU(p, q) = \frac{|p \cap g|}{|p \cup g|}
\end{eqnarray}
where $p$ and $g$ are the predicted and ground truth masks for a single mask-class.  $PQ$ is computed for each object class, and the $PQ$s for all classes are averaged as ``{\it means of PQ\/}'' ($mPQ$).  For computation of $DQ$, we use the visible mask of predicted and ground truth to find the matched instances with $IoU$ threshold of 0.5 between these predicted masks.

Since the $SQ$ was proposed as the metric of instance segmentation, which is a single mask segmentation for each object, we extended the metric to be able to evaluate multi masks segmentation for each instance as $SQ_{multi}$:
\begin{eqnarray}
  SQ_{multi} = \frac{\sum_{(p, g) \in TP} mIoU(p, q)}{|TP|} \\
  where~mIoU(p, q) = \sum_{m \in M} IoU_m(p, q) / |M|.
\end{eqnarray}
$M$ represents the set of possible instance masks, of which we have three in the instance occlusion segmentation: background, visible and occluded.  $mIoU$ (mean of $IoU$) is the averaged $IoU$ over the mask classes $m \in M$.  The $IoU$ and $mIoU$ have been used in the previous works of semantic segmentation~\cite{Everingham2014,lin2014coco} to evaluate the accuracy of predicted masks comparing it with the ground truth, so we believe the $PQ_{multi} = DQ \cdot SQ_{multi}$ is an appropriate metric for instance occlusion segmentation.  In the following we address to $PQ_{multi}$ as $PQ$ for simplicity.

%% file: src/experiments.tex
\section{EXPERIMENTS}

\subsection{Instance Occlusion Segmentation of ARC2017 Objects}

\subsubsection{Objects for Evaluation}

We evaluate our system with the 40 objects used in the Amazon Robotics Challenge 2017 (ARC2017) shown in \figref{arc2017_objects}, each of which had 4-6 instance images distributed at the competition. We believe these objects have a broad diversity and are sufficient to demonstrate picking from a pile of objects.

In the following experiments, we use the instance images distributed at ARC2017 to evaluate our image synthesis framework (\secref{image_synthesis}) comparing with human annotated data we created using the real objects. The instance occlusion segmentation model (\secref{instance_seg_nn}) is trained with the human-annotated or synthetic training dataset and is evaluated with the human-annotated testing dataset.

\begin{figure}[htbp]
  \centering
  \includegraphics[width=\columnwidth]{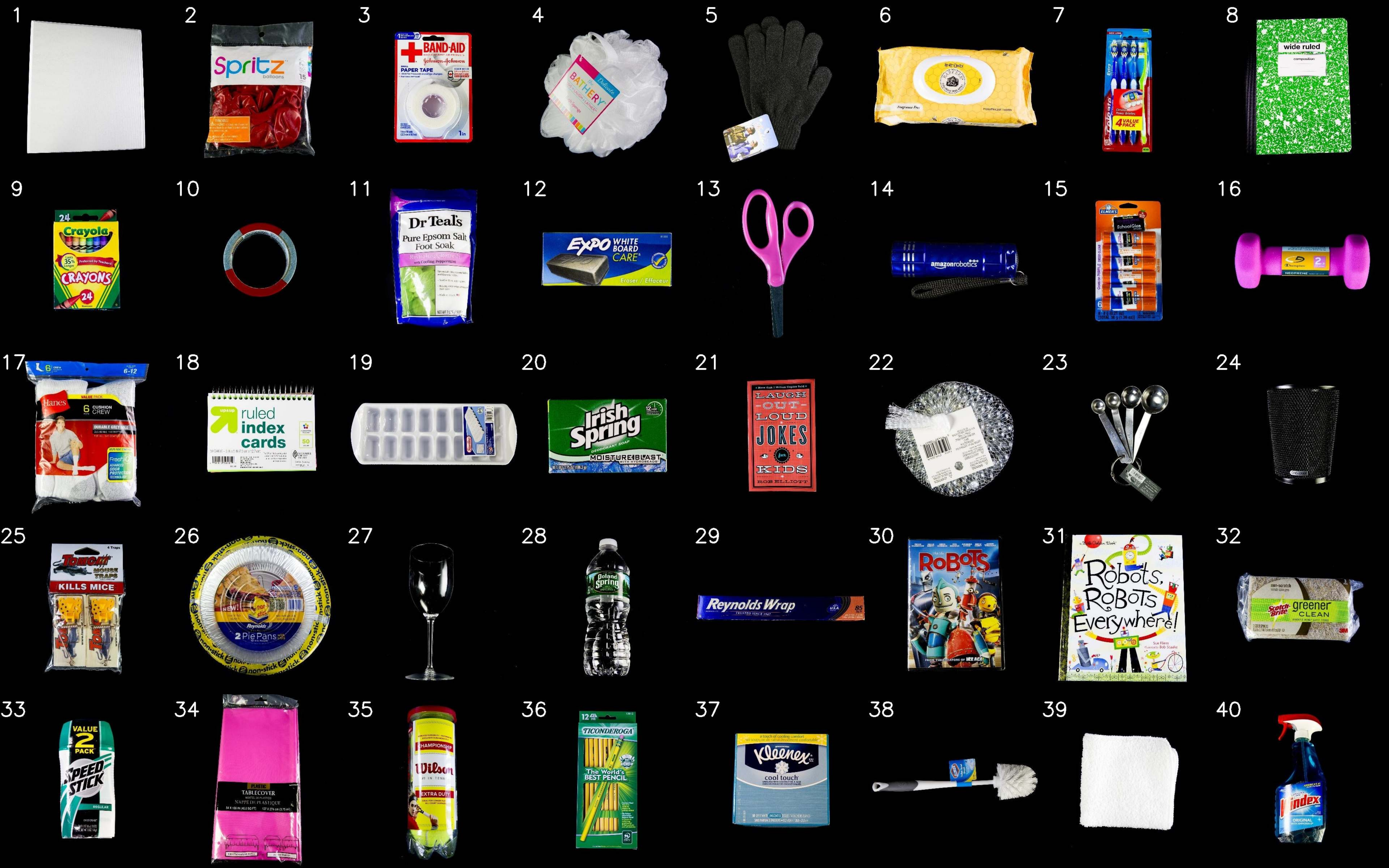}
  \caption{\textbf{40 objects used at ARC2017.}}
  \figlab{arc2017_objects}
\end{figure}


\subsubsection{Human-annotated Dataset for Evaluation}

For the evaluation of both the model and image synthesis, we created a dataset of instance occlusion segmentation shown in \figref{dataset_for_evaluation}. Since annotating an occluded region is challenging we created a set of sequential camera frames in which a pile of objects is cleared from the top. With annotating the visible mask of objects in all frames of a video captured from a fixed camera, the visible masks are backtracked to acquire the occluded masks. We created 21 videos (split into $train:test = 14:7$) in which the 40 objects appear 7 times each (3-5 times in the train split).

\begin{figure}[htbp]
  \centering
  \subfloat[Visible Masks that includes \textcolor{Green}{Baby Wipes}, \textcolor{Red}{Mouse Trap} and \textcolor{Aquamarine}{Socks}.]{
    \includegraphics[width=0.48\columnwidth]{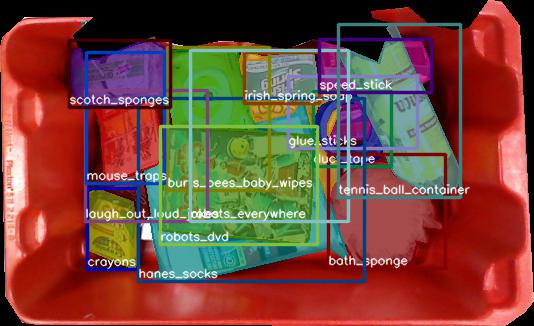}
    \figlab{dataset_for_evaluation:1}
  }
  \subfloat[Occluded mask of \textcolor{Green}{Baby Wipes}.]{
    \includegraphics[width=0.48\columnwidth]{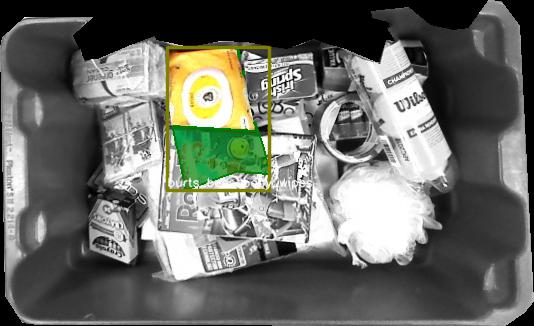}
    \figlab{dataset_for_evaluation:2}
  }

  \subfloat[Occluded mask of \textcolor{Red}{Mouse Trap}.]{
    \includegraphics[width=0.48\columnwidth]{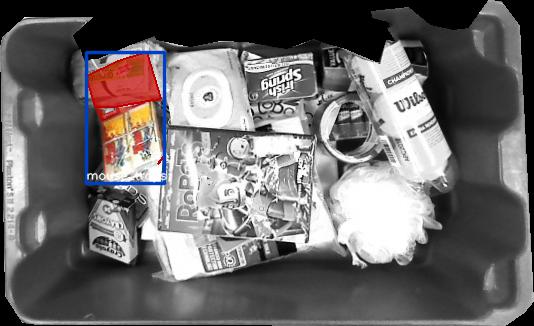}
    \figlab{dataset_for_evaluation:4}
  }
  \subfloat[Occluded mask of \textcolor{Aquamarine}{Socks}.]{
    \includegraphics[width=0.48\columnwidth]{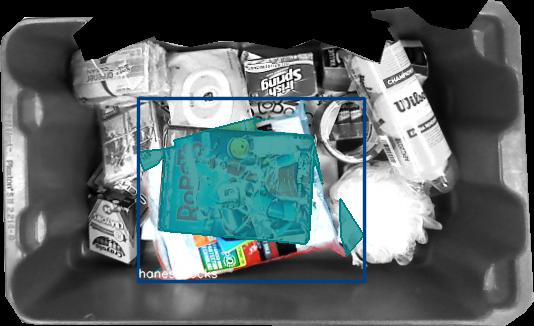}
    \figlab{dataset_for_evaluation:5}
  }
  \caption{\textbf{Annotations of instance occlusion segmentation.}\\
    \footnotesize{%
      Occluded mask of objects is visualized separately.
      Note that its color (\textcolor{Green}{Baby Wipes}, \textcolor{Red}{Mouse Trap}, \textcolor{Aquamarine}{Socks})
      corresponds to the mask of visualization of visible masks.
    }
  }
  \figlab{dataset_for_evaluation}
\end{figure}


\begin{table*}
  \begin{minipage}{\hsize}
    \centering
    \caption{\textbf{Instance occlusion segmentation on test data with ResNet101 backbone.}}
    \includegraphics[width=\textwidth]{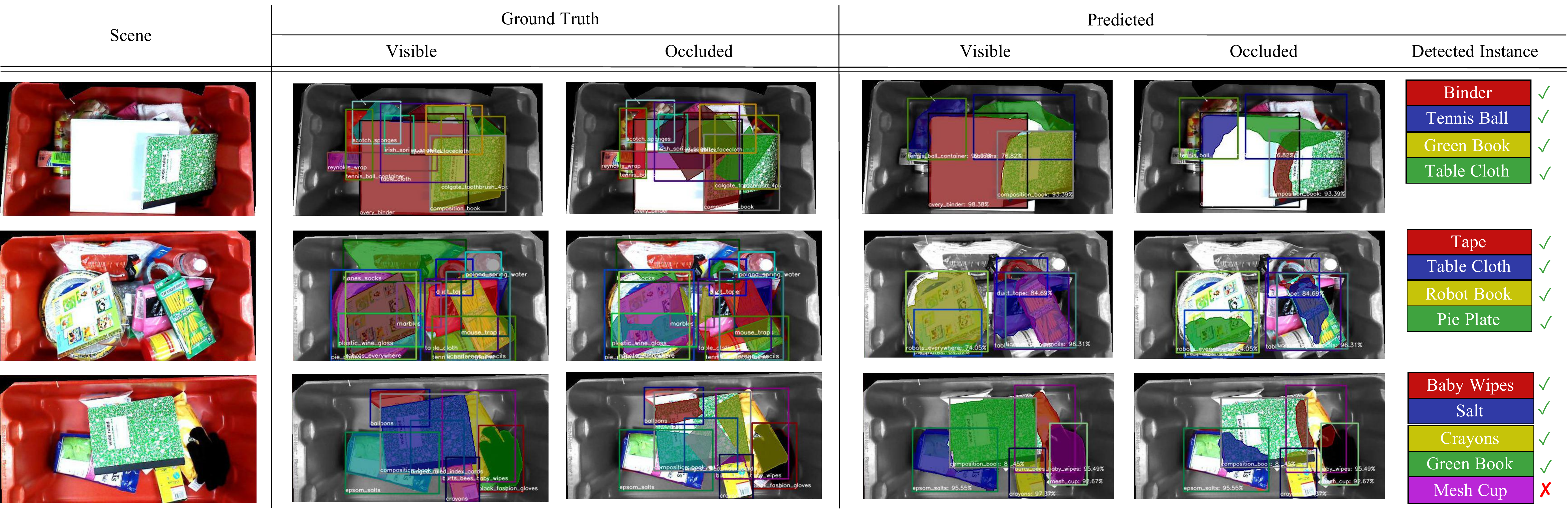}
    \tablab{qualitative}
  \end{minipage}
\end{table*}

\subsubsection{Model Evaluation: Softmax vs. Relook}\label{section:experiment:model_evaluation}

We evaluated the proposed relook architecture with a human-annotated dataset for both training and testing, with the comparison to the softmax extension of Mask R-CNN~\cite{Do2017}.  \tabref{model_evaluation} shows the quantitative results, in which ``Softmax'' denotes Mask R-CNN Softmax, ``Softmax\_x2'' denotes Softmax whose mask loss is scaled by 2, and ``Relook'' denotes Mask R-CNN with the relook architecture.  The reason why we added the experiment of ``Softmax\_x2'' is because our proposed model has two mask losses at first and second stage, possibly making it unfair to compare it with ``Softmax'', which has no scaling factor to the mask loss.  Since the learning result of RPN has lots of noise caused by randomness, we show results averaged in 10 - 15 times experiments for each model.  For the fair comparison, we use ResNet50 feature extractor using learning rate 0.0375 with 3 GPUs in which the learning rate is scaled by the number of GPUs in following experiments without note: $0.00375 = 3 \times 0.00125$, as proposed in~\cite{Goyal2017}.

The results of $mPQ$ in \tabref{model_evaluation} show that the proposed relook architecture surpasses the existing models and is effective in performing instance occlusion segmentation.  We also show results of $mAP$ (mean averaged precision) that was used as the metric of instance segmentation of visible mask in the VOC~\cite{Everingham2014} and COCO~\cite{lin2014coco} competitions.  The results of the $mAP$ show that our model surpasses previous models in both instance segmentation of visible masks and detection quality.

\begin{table}[htbp]
  \centering
  \caption{\textbf{Softmax vs. Relook.}\\
  }
  \tablab{model_evaluation}
  \begin{tabular}{c||c|ccc}
       Model & \textbf{mPQ} & mSQ & mDQ & mAP \\
       \hline
       \hline
       Softmax~\cite{Do2017} & 13.4 & 24.7 & 40.7 & 46.1 \\
       Softmax\_x2~\cite{Do2017} & 13.8 & 25.2 & 41.9 & 47.1 \\
       \textbf{Relook} (Ours) & \textbf{14.4} & \textbf{26.0} & \textbf{42.8} & \textbf{48.6} \\
  \end{tabular}
\end{table}

We also trained the network with a different backbone, ResNet101. Its qualitative results are shown in \tabref{qualitative}.


\subsubsection{Dataset Evaluation: Human-annotated vs. Synthetic}

We evaluated our image synthesis framework by training the proposed model with either synthetic or human-annotated data.  After training, the model performance is evaluated using the test split of the human-annotated dataset.  \tabref{dataset_evaluation} shows the averaged results of 3 experiments using the image synthesis referring results in \secref{experiment:model_evaluation}.  It shows that our image synthesis using 4 - 6 instance images (14.2) is as effective as the result using a small human annotated dataset (14.4) for learning instance occlusion segmentation.

\begin{table}[htbp]
  \centering
  \caption{\textbf{Human-annotated vs. Synthetic.}}
  \tablab{dataset_evaluation}
  \begin{tabular}{cc||c}
    Model & Dataset & mPQ \\
    \hline
    \hline
    \multirow{2}{*}{Softmax~\cite{Do2017}}
    & Annotated & 13.4 \\
    & Synthetic & 13.5 \\
    \hline
    \multirow{2}{*}{Relook (ours)}
    & Annotated & 14.4 \\
    & Synthetic & \textbf{14.2} \\
  \end{tabular}
\end{table}


\vspace{-5mm}
\subsection{Application to Warehouse Picking}

As an application of our system, we demonstrate the picking task of a target object from a pile of objects, as shown in \figref{demo_env}.  \figref{demo_config} shows the workspace configuration: there is a bin which contains the target object, a bin to place obstacle objects into it, and a cardboard box for the robot to place the target object.  For this demonstration, we have extended the picking task we developed in the previous works~\cite{wada2017probabilistic,Hasegawa2017}.  \figref{picking_demo} shows the sequential frames of recognition result (\figref{picking_demo_rviz:1}-\ref{figure:picking_demo_rviz:3}), workspace overview (\figref{picking_demo:1}-\ref{figure:picking_demo:6}) and emphasis on the robot hand (\figref{picking_demo_zoom:1}-\ref{figure:picking_demo_zoom:6}).  The recognition result shows the input image, predicted visible and occluded masks, and the target object based on the occlusion understanding.

In this experiment, we set the threshold of occlusion ratio to 0.3; the occlusion ratio is the ratio of occluded pixels compared with the total pixels of an instance.  We set the threshold of inter-instance occlusion ratio to 0.1 to judge that the instance is occluded by the other.  For the model of this demonstration, we use ReNet101 feature extractor as the backbone of our proposed model to have a better recognition accuracy.  The successful clearing of obstacle objects and picking targets in the demonstration shows that our proposed system is effective and applicable in the real-world picking task.

\begin{figure}[htbp]
  \centering
  \subfloat[The Scene of Stacked Objects.]{
    \includegraphics[width=0.57\columnwidth]{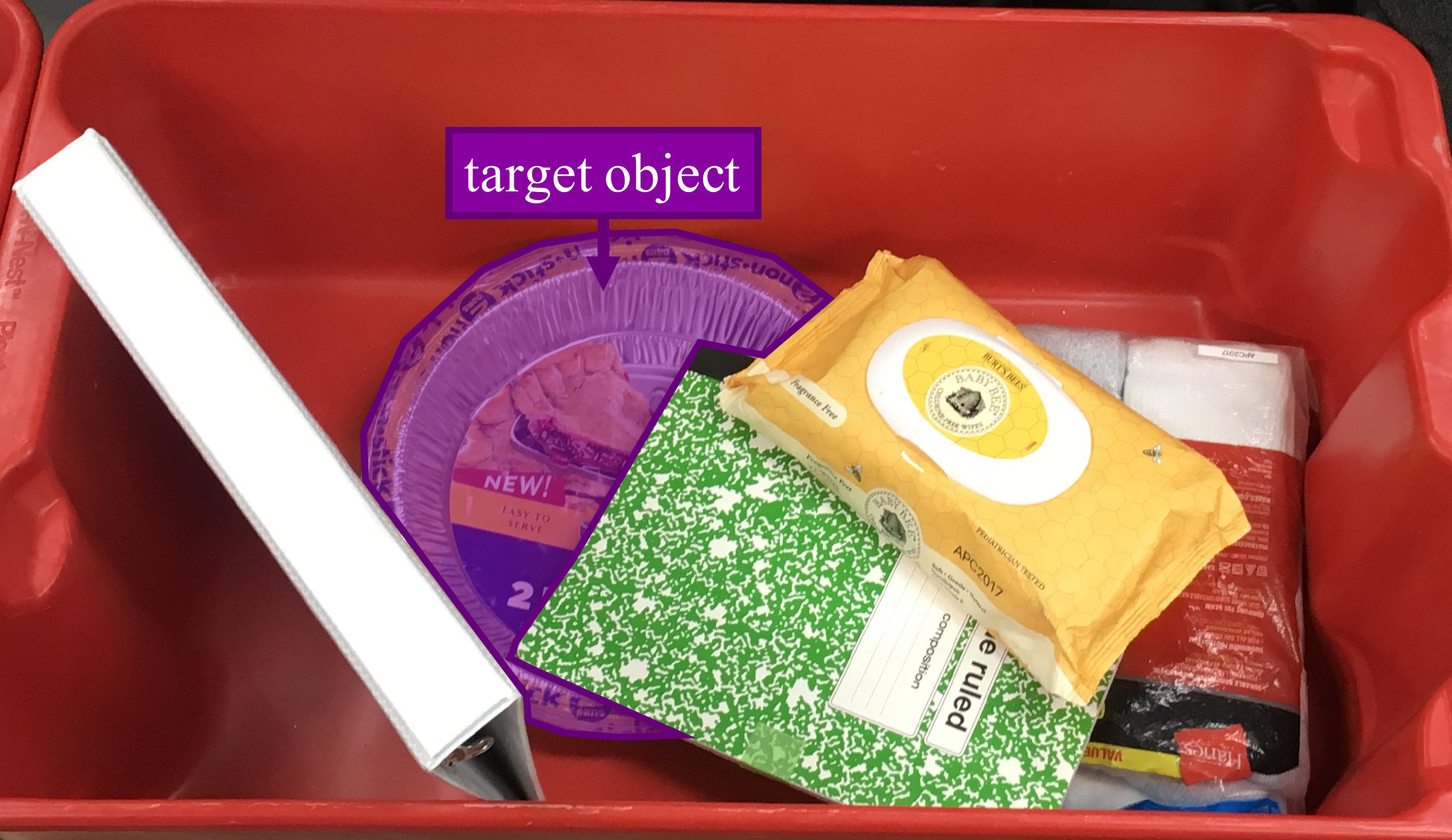}
    \figlab{demo_env}
  }
  \subfloat[Workspace.]{
    \includegraphics[width=0.41\columnwidth]{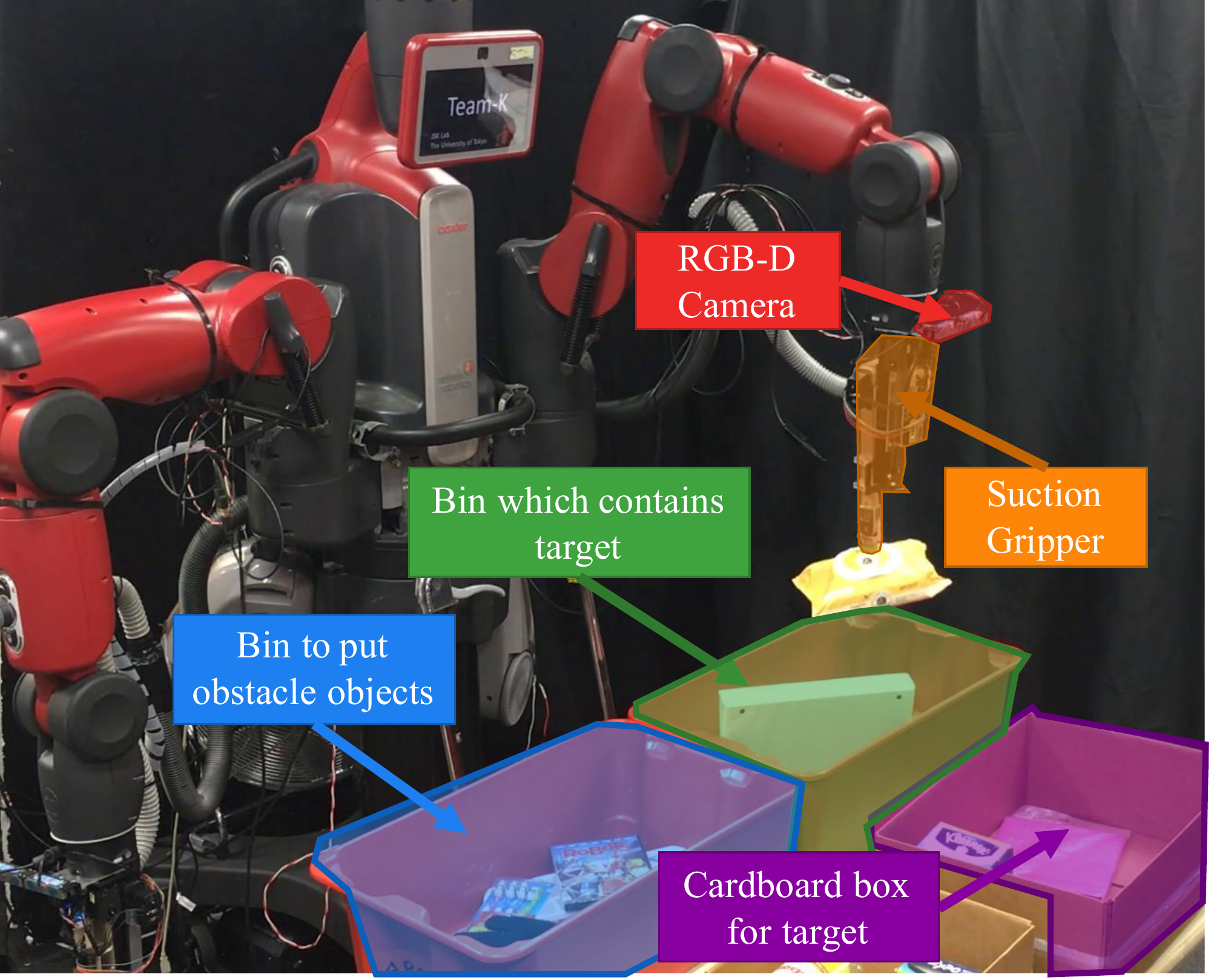}
    \figlab{demo_config}
  }
  \caption{\textbf{Picking Demo Configuration.}}
\end{figure}

\begin{figure*}[htbp]
  \centering
  \subfloat[]{
    \includegraphics[width=0.315\textwidth]{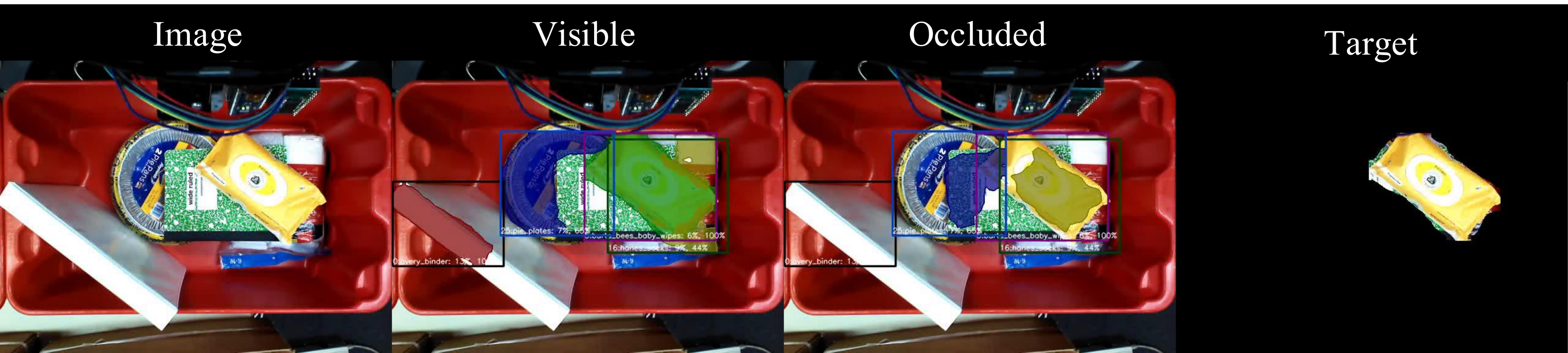}
    \figlab{picking_demo_rviz:1}
  }
  \subfloat[]{
    \includegraphics[width=0.315\textwidth]{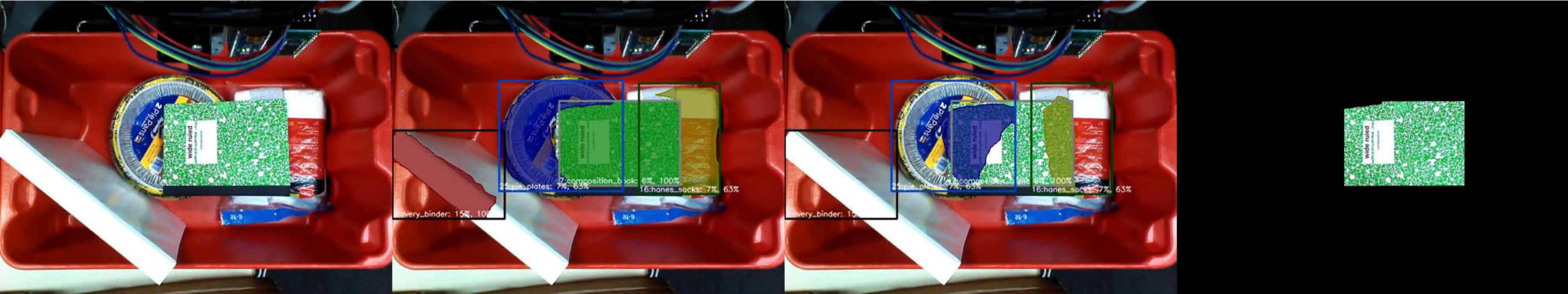}
    \figlab{picking_demo_rviz:2}
  }
  \subfloat[]{
    \includegraphics[width=0.315\textwidth]{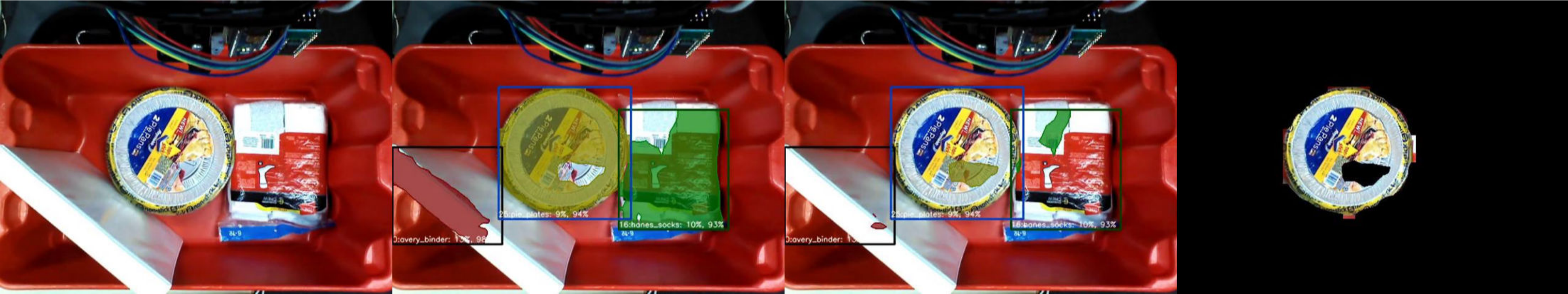}
    \figlab{picking_demo_rviz:3}
  }

  \subfloat[]{
    \includegraphics[width=0.31\columnwidth]{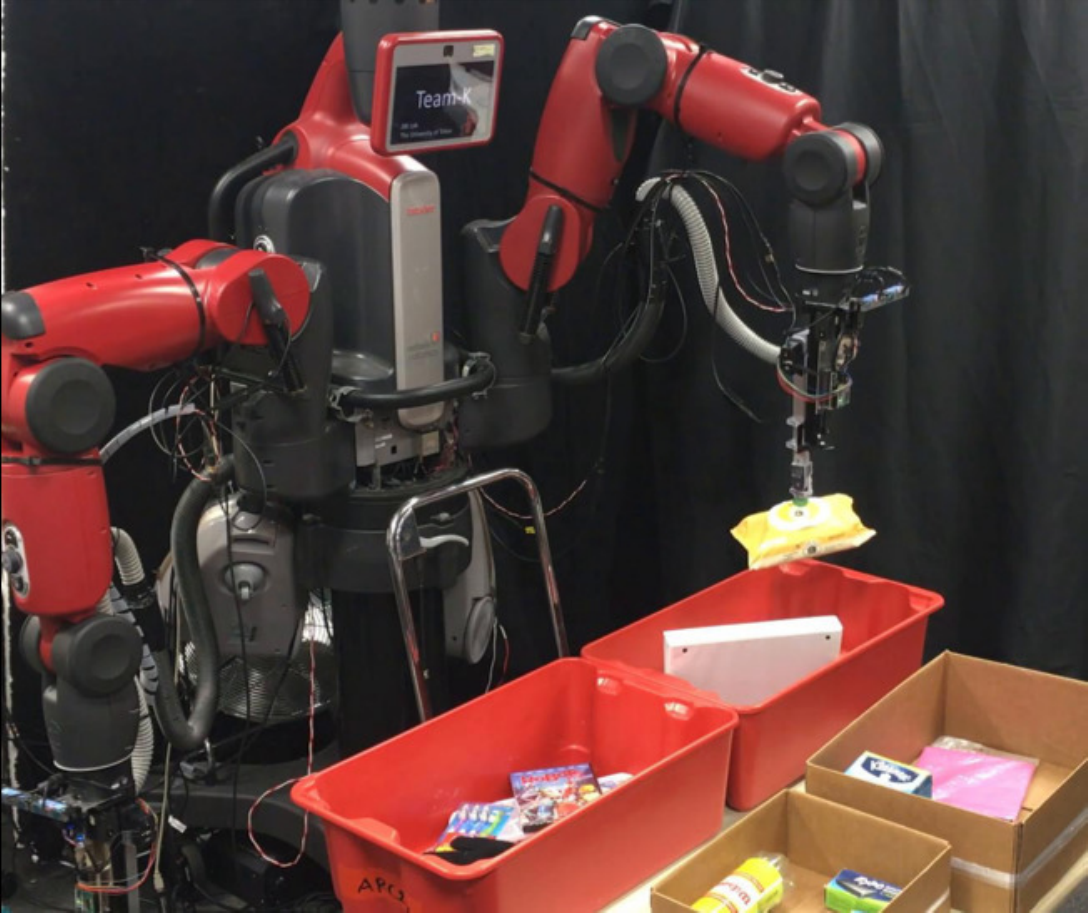}
    \figlab{picking_demo:1}
  }
  \subfloat[]{
    \includegraphics[width=0.31\columnwidth]{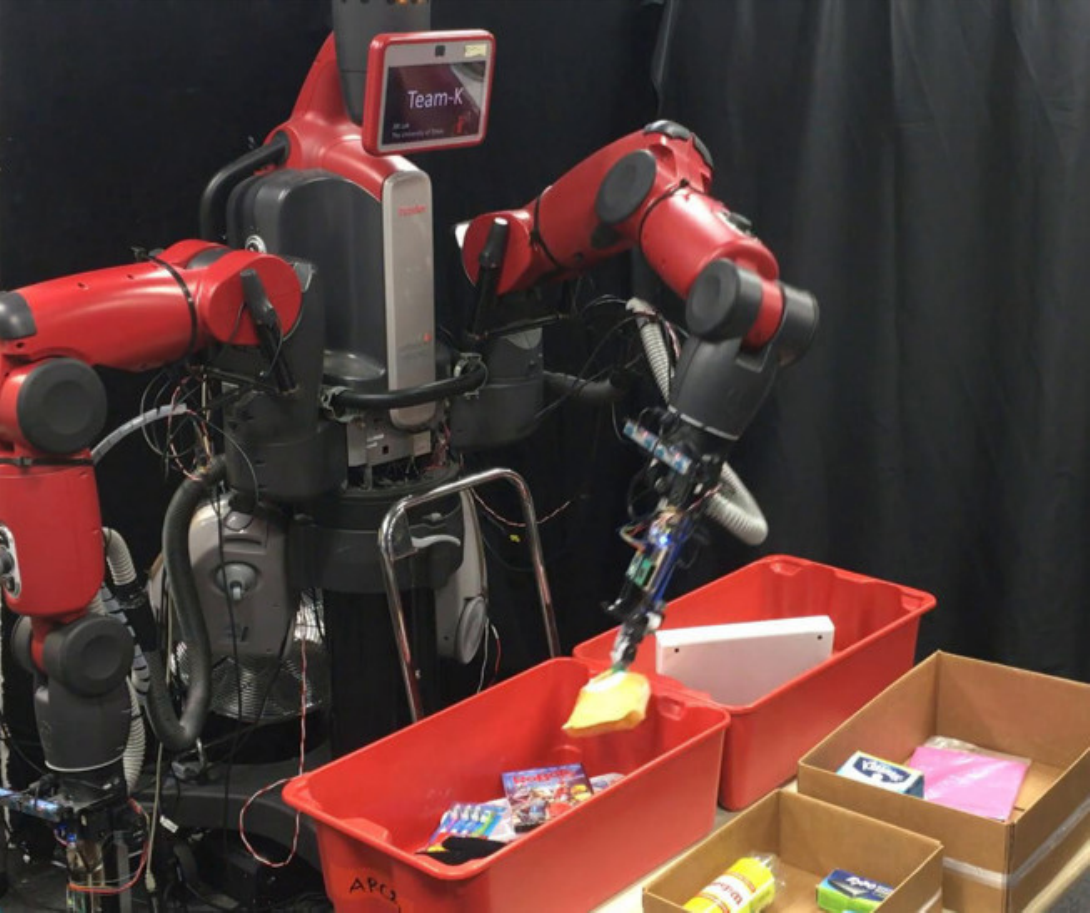}
    \figlab{picking_demo:2}
  }
  \subfloat[]{
    \includegraphics[width=0.31\columnwidth]{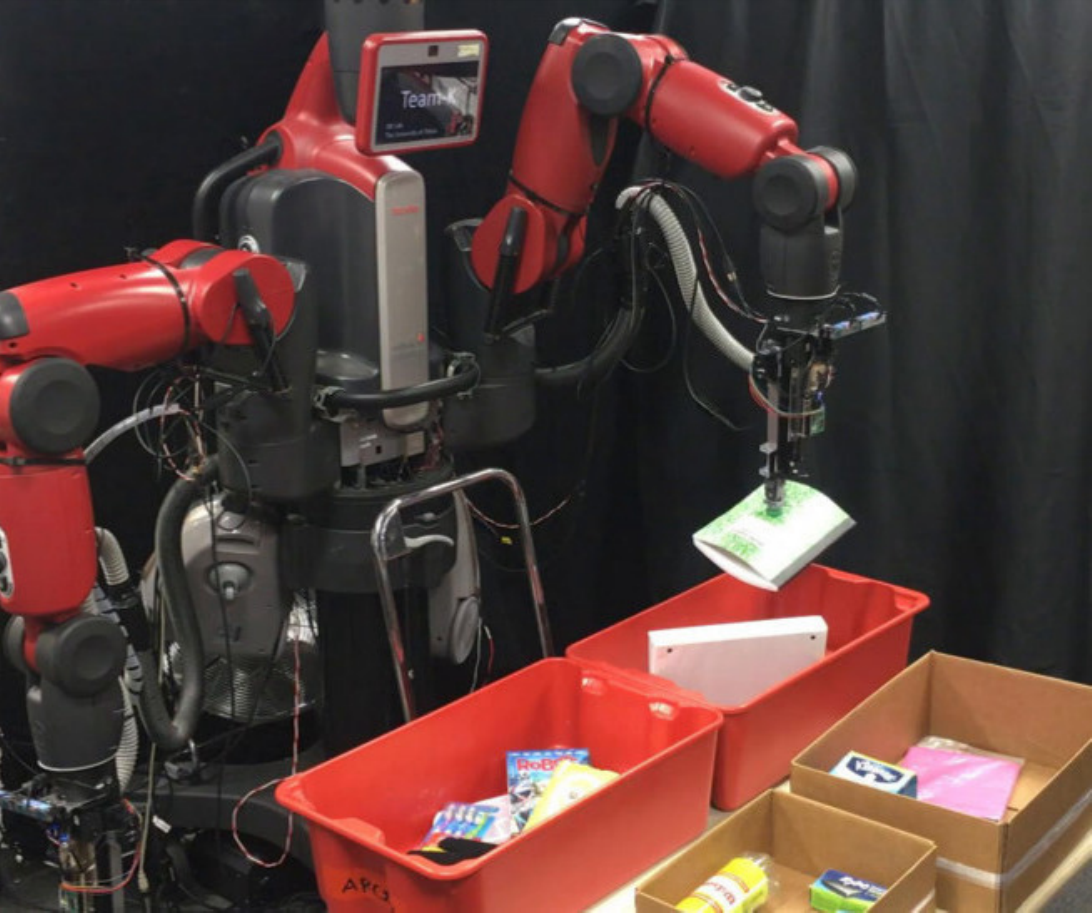}
    \figlab{picking_demo:3}
  }
  \subfloat[]{
    \includegraphics[width=0.31\columnwidth]{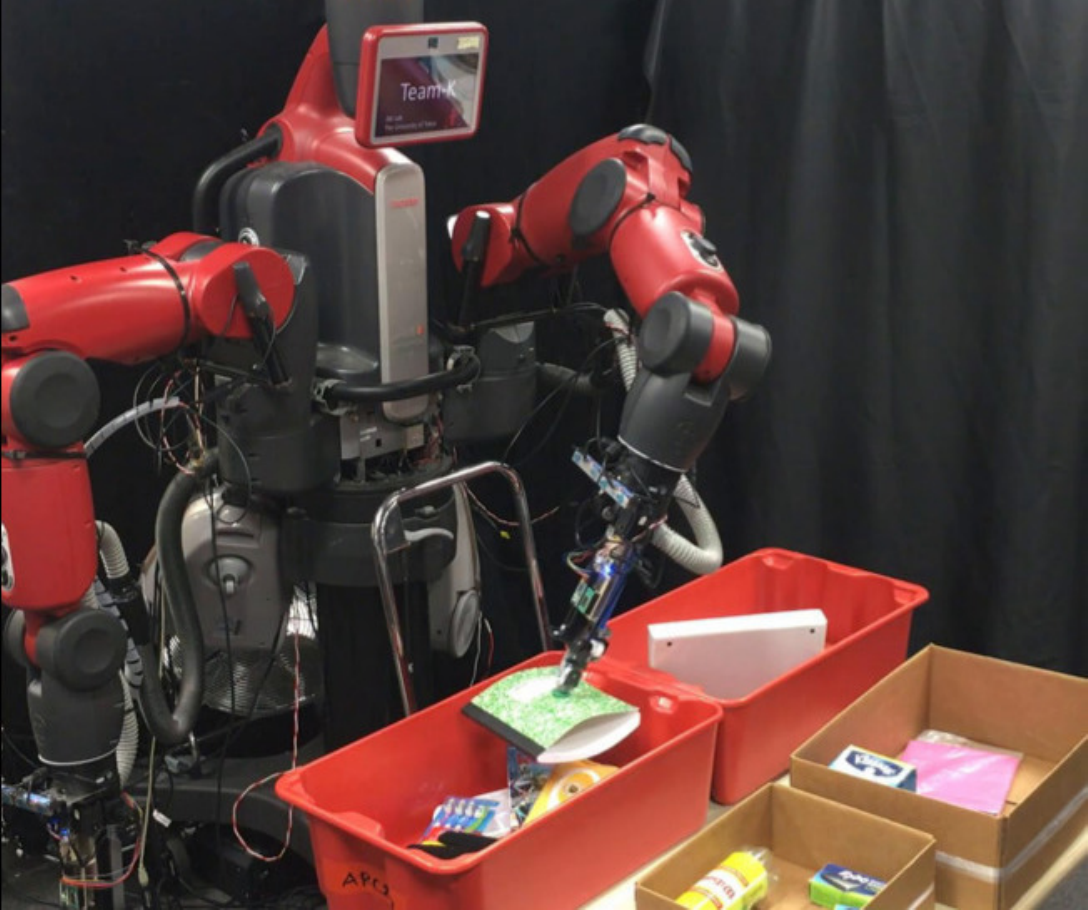}
    \figlab{picking_demo:4}
  }
  \subfloat[]{
    \includegraphics[width=0.31\columnwidth]{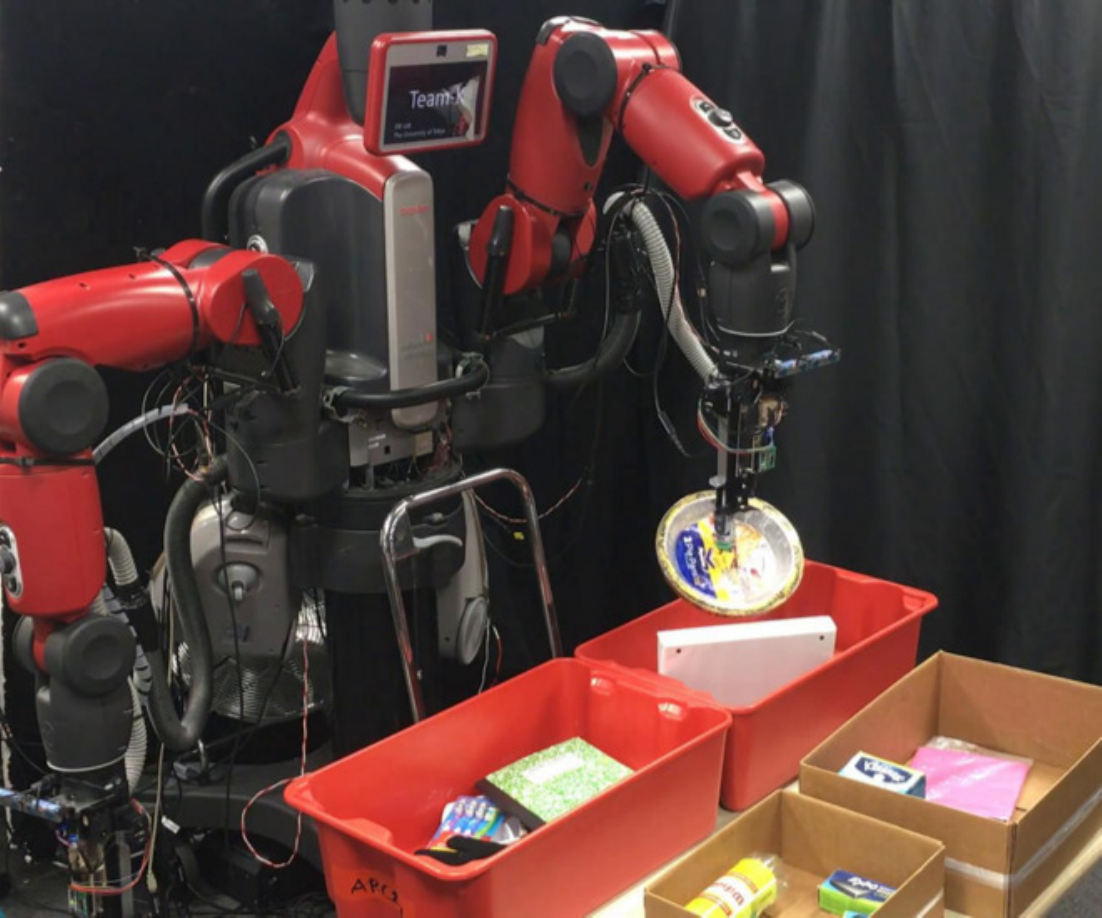}
    \figlab{picking_demo:5}
  }
  \subfloat[]{
    \includegraphics[width=0.31\columnwidth]{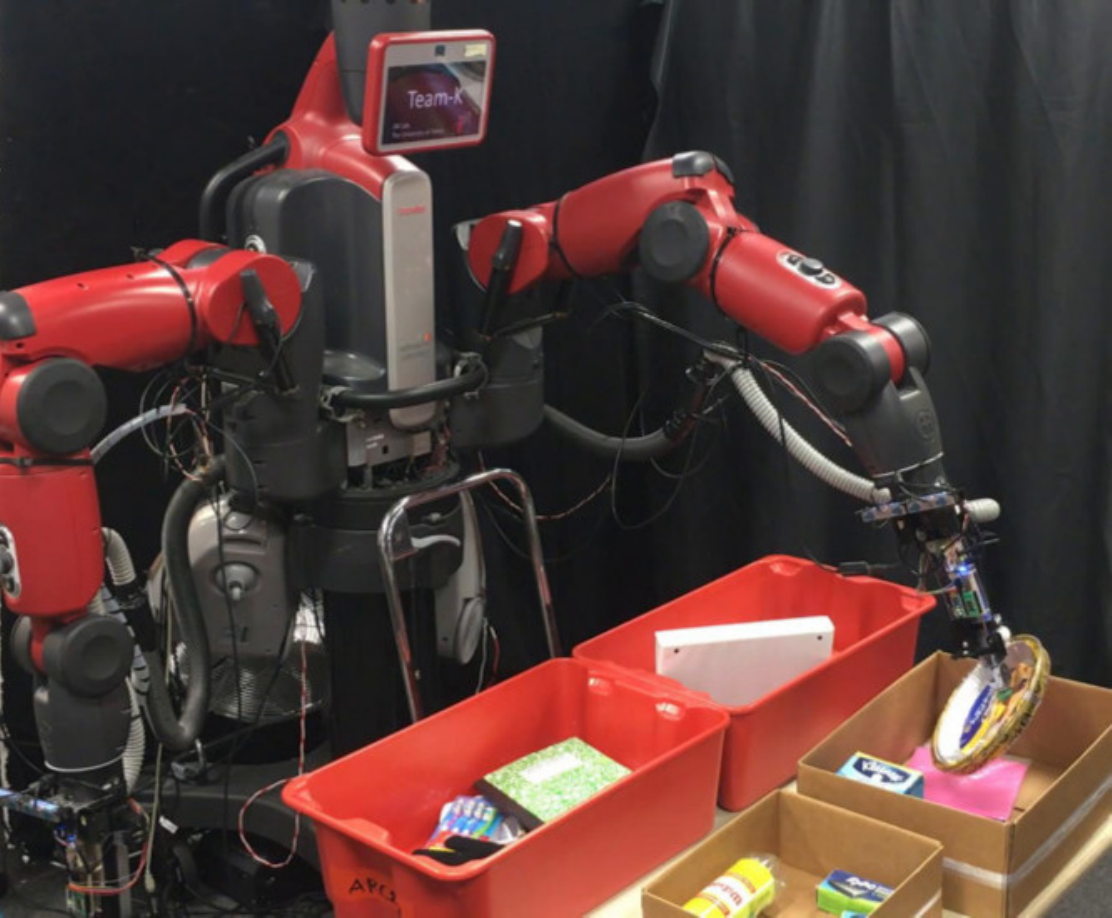}
    \figlab{picking_demo:6}
  }

  \subfloat[]{
    \includegraphics[width=0.31\columnwidth]{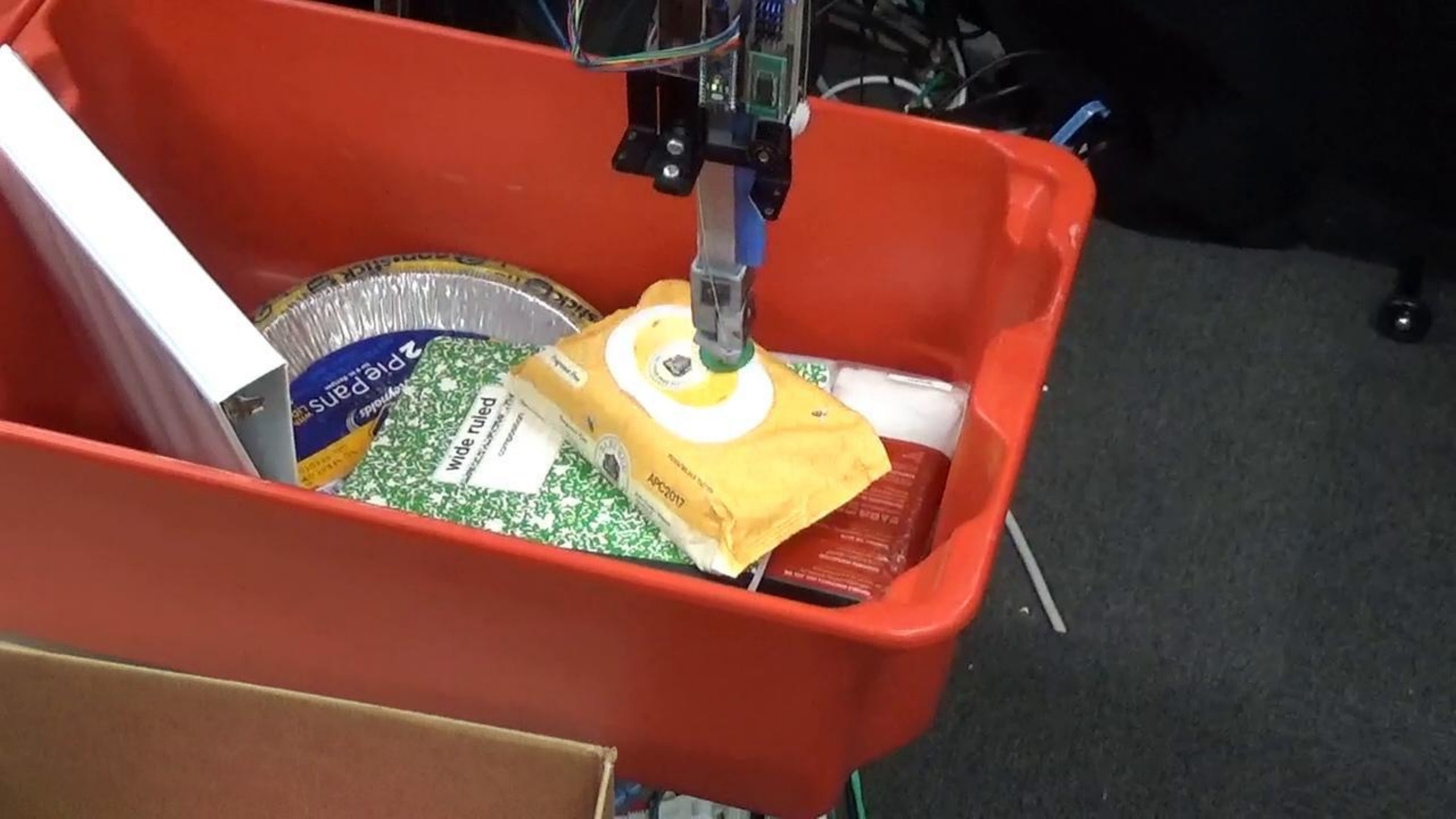}
    \figlab{picking_demo_zoom:1}
  }
  \subfloat[]{
    \includegraphics[width=0.31\columnwidth]{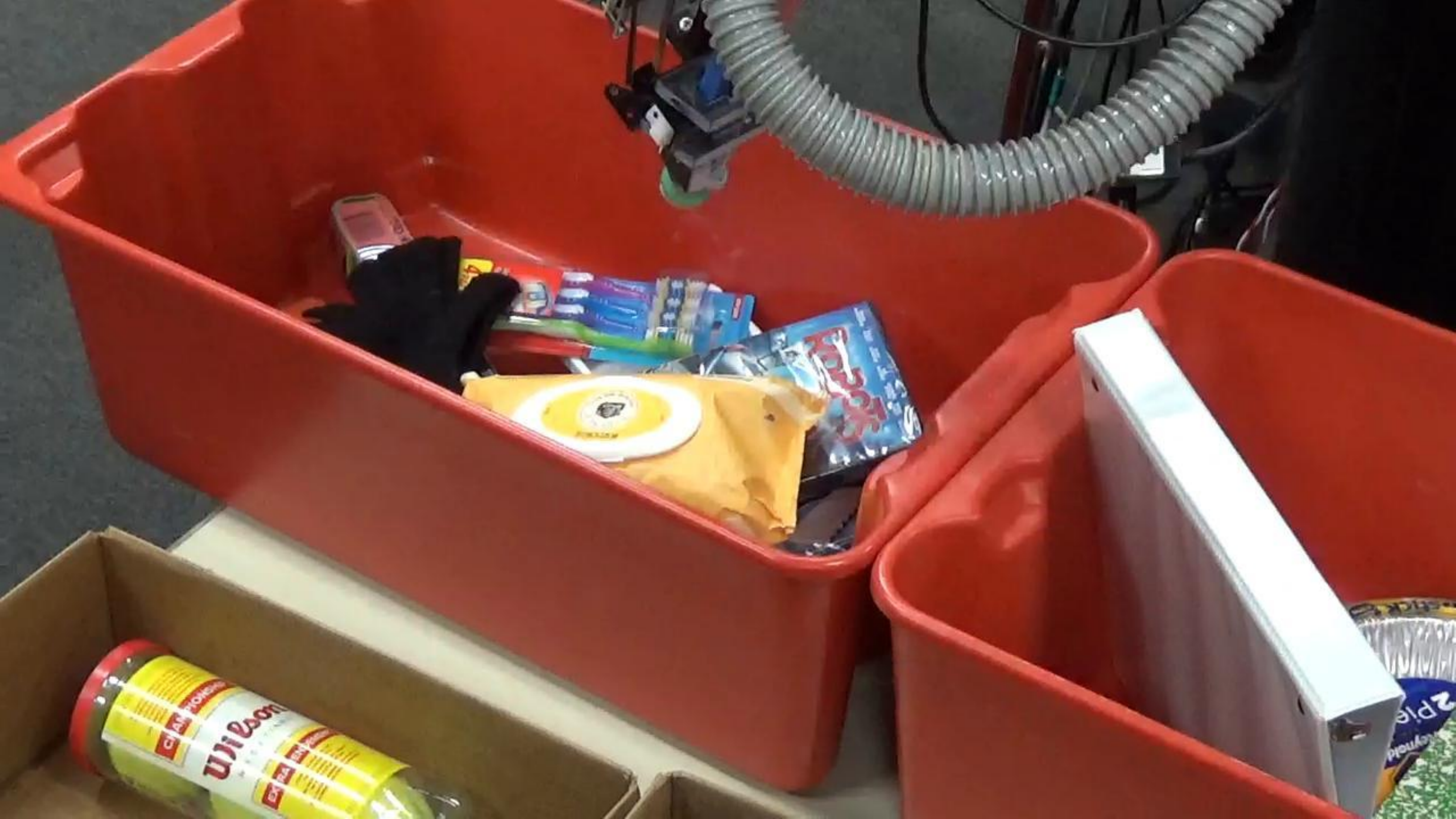}
    \figlab{picking_demo_zoom:2}
  }
  \subfloat[]{
    \includegraphics[width=0.31\columnwidth]{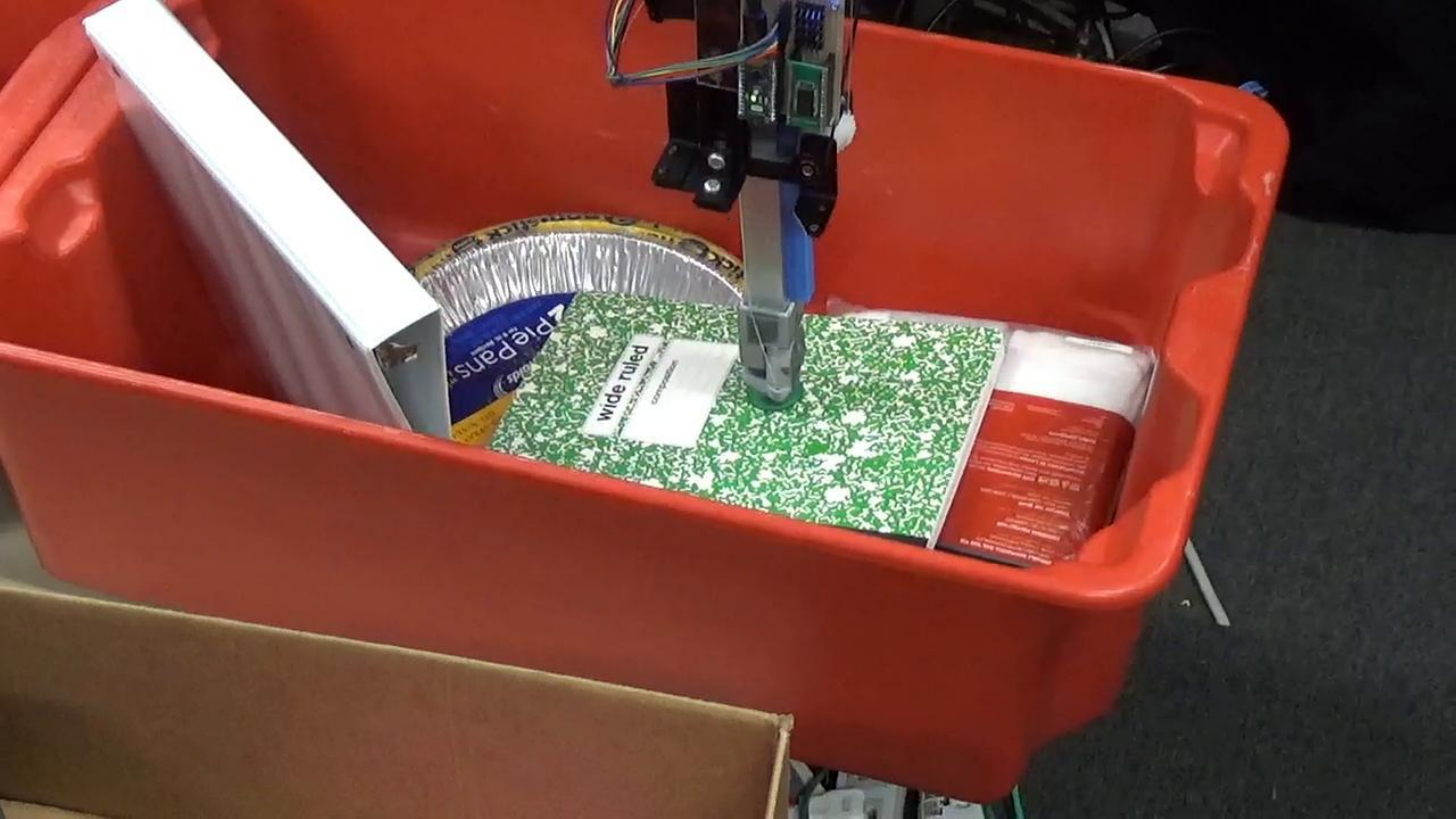}
    \figlab{picking_demo_zoom:3}
  }
  \subfloat[]{
    \includegraphics[width=0.31\columnwidth]{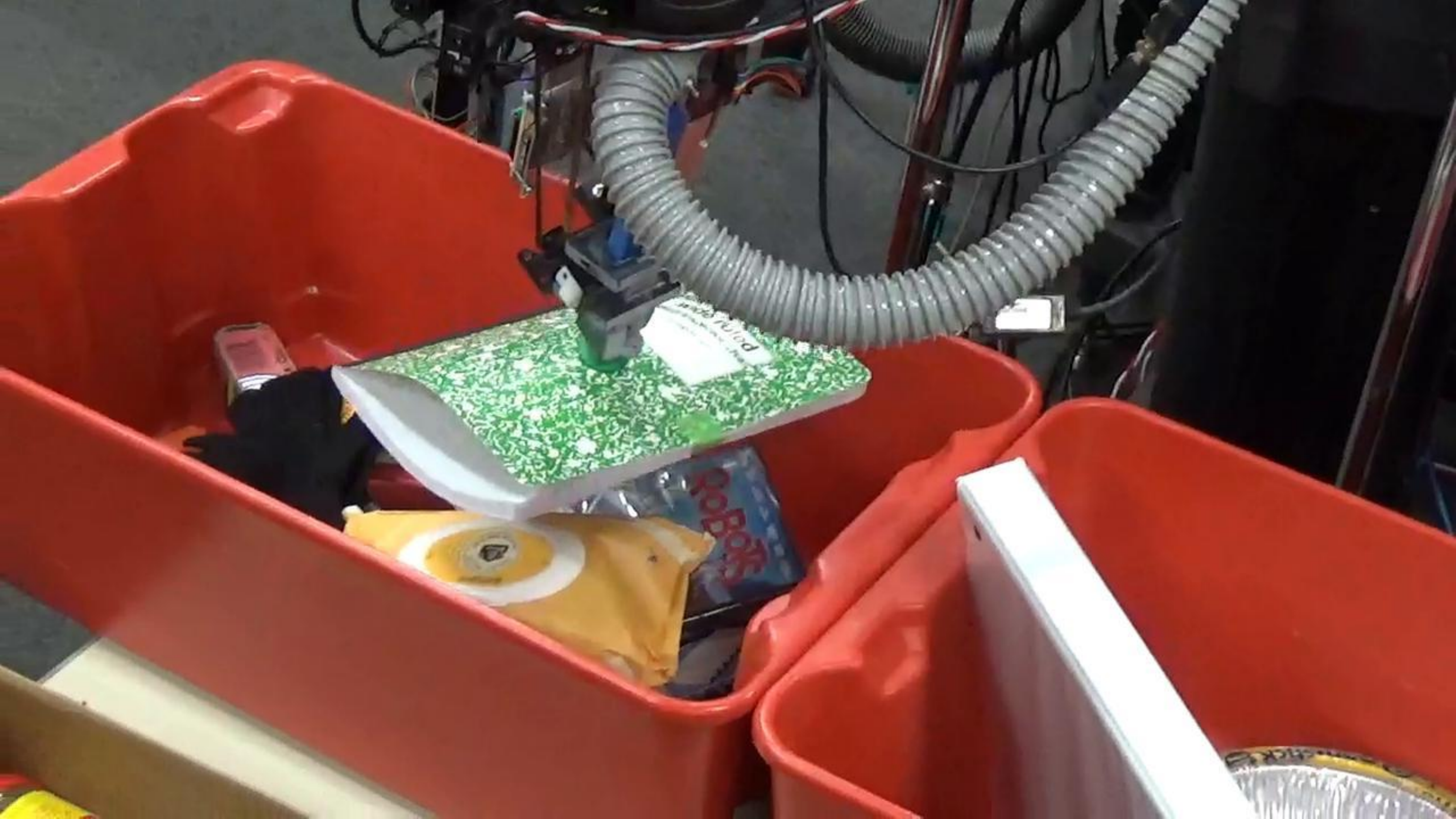}
    \figlab{picking_demo_zoom:4}
  }
  \subfloat[]{
    \includegraphics[width=0.31\columnwidth]{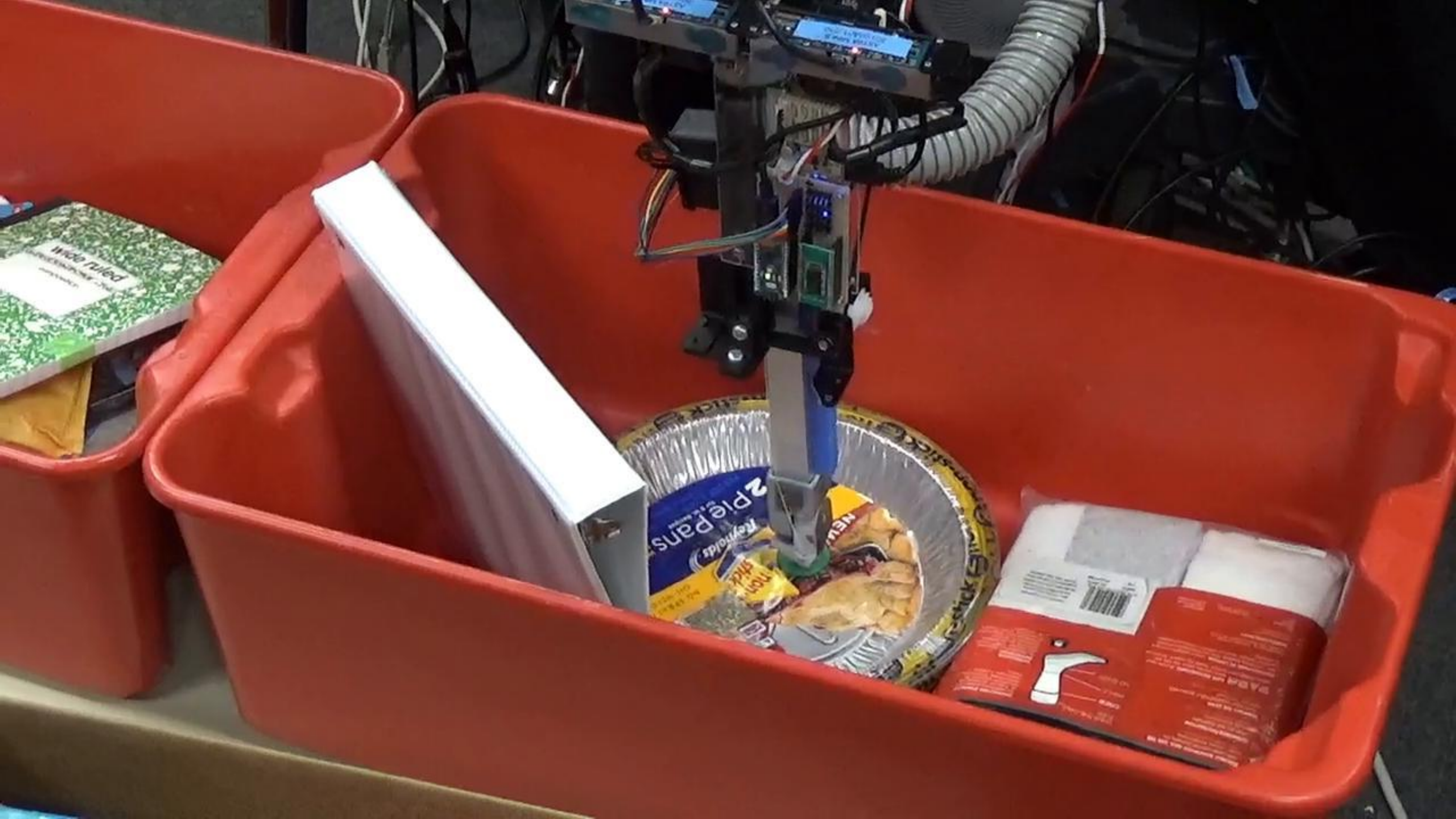}
    \figlab{picking_demo_zoom:5}
  }
  \subfloat[]{
    \includegraphics[width=0.31\columnwidth]{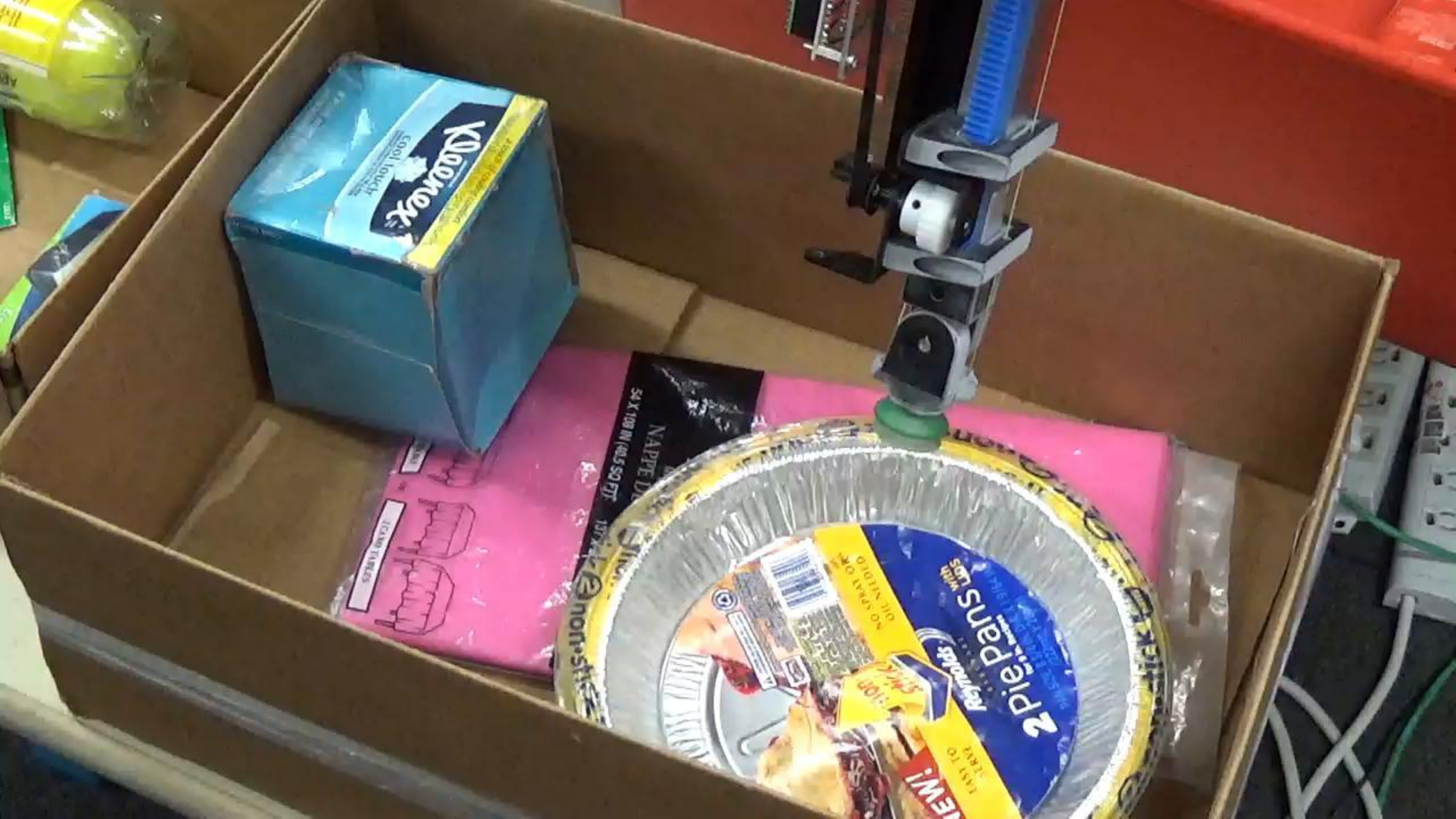}
    \figlab{picking_demo_zoom:6}
  }
  \caption{\textbf{Picking Task Demonstration of a Target from a Pile of Objects.}}
  \figlab{picking_demo}
\end{figure*}

%% file: src/conclusions.tex
\section{CONCLUSIONS}

We presented a vision system that only requires a few instance images of objects to learn instance occlusion segmentation. The system consists of 1) Image synthesis with ground truth of occluded region mask of each instance; 2) Instance segmentation networks that learn inter-instance relationship, which is essential for the segmentation of occluded regions. We evaluated the proposed image synthesis and segmentation model via the ablation studies and presented the effectiveness of the proposed system in the real picking task from a pile of objects.

%% file: src/acknowledgement.tex
\section*{ACKNOWLEDGEMENT}

We thank Shun Hasegawa and Yuto Uchimi for contributions
to the software and hardware development of the picking system for demonstration
with real-world robot experiments.